\documentclass[12pt,final]{elsarticle}
\usepackage[top=2.5cm,bottom=2.5cm,left=2.5cm,right=2.5cm]{geometry}
\usepackage[colorlinks=true,  linkcolor=blue, citecolor=blue]{hyperref}
\usepackage{commath}
\usepackage{caption}
\usepackage{amsmath}
\usepackage{amssymb}
\usepackage{bm}
\usepackage{graphicx}
\usepackage{subfig}
\usepackage{tabularx}
\usepackage[inline]{enumitem}
\usepackage{amsfonts}
\usepackage{varwidth}
\usepackage{cleveref}
\usepackage[vlined]{algorithm2e}
\usepackage{setspace,kantlipsum}

\usepackage{alphalph}

\usepackage{endnotes}

\journal{Signal Processing}
\date{24 June 2020}
\begin{document}

\begin{frontmatter}

\title{Fast and robust small infrared   target detection using absolute directional   mean difference algorithm}

\author{Saed Moradi}
\author{Payman Moallem\corref{colabel}}
\ead{p\_moallem@eng.ui.ac.ir}
\author{Mohamad Farzan Sabahi\corref{}}
\address{Department of Electrical Engineering, Faculty of Engineering, University of Isfahan, Isfahan, Iran}
\cortext[colabel]{Corresponding author.}
\begin{abstract}
Infrared small target detection in an infrared  search and track (IRST) system  is a challenging task. This situation becomes more complicated when high gray-intensity structural backgrounds appear in the field of view (FoV) of the infrared seeker. While the majority of the infrared small target detection algorithms neglect directional information, in this paper, a directional approach is presented to suppress structural backgrounds and develop a more effective detection algorithm. To this end,  a similar concept to the average absolute gray difference (AAGD) is utilized to construct a novel directional small target detection algorithm called absolute directional  mean difference (ADMD). Also, an efficient implementation procedure is presented for the proposed algorithm. The proposed algorithm effectively enhances the target area and eliminates background clutter. Simulation results on real infrared images prove the significant effectiveness of the proposed algorithm.

\end{abstract}


\end{frontmatter}
\section{Introduction}
Infrared imaging plays a vital role in a  wide range of  applications in remote sensing \cite{turker2017review,jin2011evaluation} and long-distance target detection is of great importance for surveillance applications. In this filed, small infrared target detection remains  a major challenge despite  advances in infrared detector technology and image processing techniques \cite{liu2018infrared}. This challenge becomes more critical when the long-distant target is embedded in a noisy and complicated background \cite{khaledian2014new}.

There are too many researches in the literature which are dedicated to the small infrared target detection field. The target detection procedure can be considered as a binary classification problem. Therefore, the detection process can be done using machine learning-based algorithms and different classifiers. Principal component analysis (PCA) \cite{hu2010infrared}, probabilistic principal component analysis (PPCA) \cite{wang2015adaptive,cao2008infrared},  robust principal component analysis (RPCA) \cite{dai2016infrared}, and nonlinear principal component analysis (NPCA) \cite{liu2005detection} were presented for small infrared target detection. Sparse representation-based methods detect small targets by decomposing each image patch over an over-complete dictionary. The over-complete dictionary can be constructed using Gaussian spatial distribution model for the small infrared target \cite{zhao2011infrared}. Also, the dictionary may be constructed according to background features \cite{liu2017infrared}. Support vector machine (SVM) \cite{bi2017multiple}, and artificial neural networks \cite{shirvaikar1995neural} are also used as  classifiers to distinguish  small targets from background clutters. While the machine learning-based methods may work well in some situations, the overall detection ability and false-alarm rate  drastically depend on the training samples and features. 

Single-frame filtering-based methods are the most practical methods in the literature. These methods are divided into two categories. The first category is  background subtraction. The algorithms belong to this category like as max-mean \cite{deshpande1999max}, max-median \cite{deshpande1999max}, and morphology opening \cite{zeng2006design} are  fast and  have low  computational complexity. Although, these methods have a high false-alarm rate due to imperfect background estimation.  The other type of single-frame filtering-based small target detection algorithms simultaneously enhances the target area and suppresses the background clutter. Laplacian of Gaussian (LoG) scale-space was proposed to distinguish the target from background clutter by maximum selection along scale dimension \cite{kim2012scale}. An extension to this method was presented in \cite{shao2012improved}, where the morphological operators were utilized after LoG filtering to eliminate remaining  background clutter. While the LoG operator  is able to detect low-contrast point-wise targets, the second derivative operator boosts background noise. The difference of Gaussian (DoG) which is an efficient approximation of LoG was used in \cite{wang2012infrared,dong2014infrared} for detecting dim and small targets. Difference of Gabor (DoGb) and improved difference of Gabor (IDoGb) were presented in \cite{han2016infrared} to address false responses of the DoG filter along edges. The combination of DoG and frequency domain approach was presented in \cite{wang2017infrared}. Local contrast measure (LCM) \cite{chen2014local} has a good ability to enhance target region. However, LCM amplifies the single-pixel salt noise and increases the number of false-alarms. Additionally, the high-intensity background clutters are also intensified in the output image. Improved local contrast measure (ILCM) \cite{han2014robust}, novel local contrast measure (NLCM) \cite{qin2016effective}, and relative local contrast measure (RLCM)       \cite{han2018infrared} are the extended algorithms inspired by human visual system. 
Average absolute gray difference  (AAGD) algorithm \cite{deng2016infrared} effectively suppresses background noise (using local averaging) and enhances target area (by adopting local contrast). Since this algorithm can be easily implemented via local averaging and subtraction operators, the AAGD algorithm  has low computational complexity  and is suitable for real-time practical applications. However, AAGD faces a major problem when the infrared scenario contains high-intensity edges and structural background clutter. In this situation, AAGD returns false responses as true target areas.
In multi-scale patch-based contrast measure (PCM) \cite{wei2016multiscale}, after dividing each image patch into nine cells, the dissimilarities between the surrounding  cells and the central one are calculated. By multiplying directional dissimilarities  and minimum selection among the results, the final output is obtained. While this method utilizes the directional approach, however, by multiplying the directional dissimilarities, the final saliency map contains strong responses in non-target area. For instance, directional dissimilarities have high negative value for hole-like small objects. Thus, multiplying these negative values leads to high-intensity false responses.   Note that, multiplying different directional information is the mainly used  directional data combination in the literature   \cite{aghaziyarati2019small,zhang2018directional,zhang2017infrared,qi2012robust,yang2014multiscale}.

 The temporal information through successive frames can also be used to further false-alarms reduction \cite{zhao2018spatial}. Generally, in spatio-temporal (multi-frame) methods,  the moving small infrared target is distinguished from random noise and cloud clutter using a  predefined motion model \cite{chen2014novel}. However, there are two main challenges regarding utilizing multi-frame methods. First, the successive frames should be registered before processing, while  the registration procedure is a challenging task for the textureless sky background. Second, the target with radial motion cannot be identified using this approach because the location of the target does not change over the successive frames.

In order to develop a robust and effective small target detection algorithm and address the aforementioned issues, in this paper, a directional approach is presented and a novel algorithm called absolute directional mean difference (ADMD) is proposed. The rest of this paper is organized as follows. In the next section, the related background and  motivation is discussed briefly. The third section  is dedicated to the proposed small target detection algorithm followed by simulation results in the fourth section. Finally, the paper is concluded in the fifth section.

\section{Background and motivation}
\subsection{Average absolute gray difference algorithm}
\begin{figure}[!t]
\centering
\includegraphics[width=0.25\textwidth]{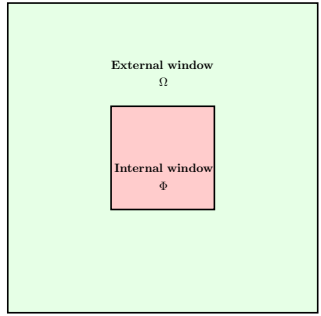}
\caption{The internal  as well as the external windows in the AAGD algorithm}
\label{fig:aagd_win}
\end{figure}

Average absolute gray difference (AAGD) is a recently presented algorithm for small infrared target detection in the literature  \cite{deng2016infrared}. Since this algorithm utilizes local averaging, it has good capability to eliminate background noise. The target enhancement mechanism in this method relies on the existing contrast between the target region and local background area.  This algorithm is utilizing two nested windows which are slid pixel by pixel.  The difference between the average values of the internal window $\Phi$ (\autoref{fig:aagd_win}) and the external window $\Omega$  is used to construct a saliency map. Note that, the internal and external windows may be called target and background windows, respectively. The saliency map construction in AAGD algorithm can be mathematically formulated as
\begin{equation}
AAGD=\abs{\frac{1}{N_{\Phi}}\sum_{(s,t)\in\Phi}{I(s,t)}-\frac{1}{N_{\Omega}}\sum_{(p,q)\in\Omega}{I(p,q)}}^2 ,
\label{eq:aagd}
\end{equation}
where $I(x,y)$, $N_{\Phi}$, and $N_{\Omega}$ stand for the pixel intensity at position $(x,y)$, the total number of pixels contained in the set $\Phi$, and the total number of pixels contained in the set $\Omega$, respectively.

AAGD algorithm has a good ability to enhance low contrast target and eliminate background noise. However, when the infrared scenario contains high-intensity edges and structural background clutter, non-target areas are also intensified. Also, while the target area  always has positive contrast, AAGD returns strong responses no matter whether the contrast of an interested region is positive or negative  \cite{moradi2018false}.

To address the aforementioned issues and overcome deficiencies of the AAGD algorithm, here, a directional approach is utilized  to achieve better clutter rejection and background suppression capability. The proposed approach is adopted in the AAGD algorithm and a new small infrared target detection algorithm called  absolute directional  mean difference (ADMD) is developed. The following section describes the proposed algorithm.

\section{The proposed algorithm}
While the directional filter design concept was presented and investigated  in small infrared target detection field  \cite{Qi2013robust,yang2015directional,wei2016multiscale,zhang2017infrared}, the lack of proper directional data combination strategy avoids robust small target detection. As mentioned in the introduction section, multiplication is the  mainly used  data combination strategy that has been investigated in the literature. Using multiplication to obtain final output from directional filters (similar to what proposed by \cite{Qi2013robust,yang2015directional,wei2016multiscale,zhang2017infrared}) will intensify the non-target areas. In order to address this issue and avoiding information loss, the combination strategy is modified in a way to improve clutter rejection ability. 

\begin{figure}[!t]
\centering
\includegraphics[width=0.5\textwidth]{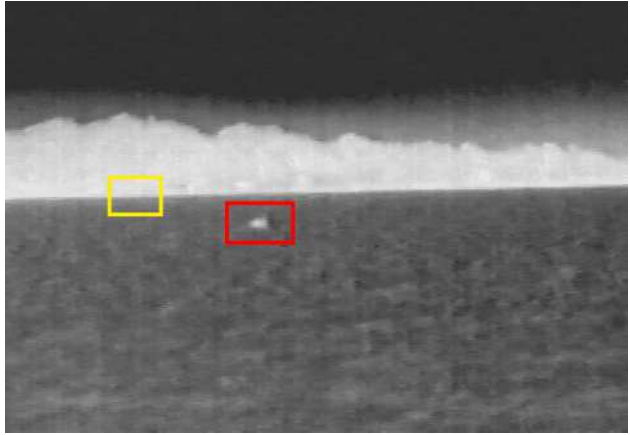}
\caption{A target and a high-intensity edge in real infrared scenario}
\label{fig:tar_edge}
\end{figure}

Generally, infrared targets have  positive local contrast which means that the target areas  are brighter than the local background in all directions (neglecting special cases like the situation which the target is close to high-intensity background clutter). Considering this fact into account, it appears that utilizing all information from the all directional filters effectively can suppress  high-intensity structural  background. As shown in \autoref{fig:tar_edge}, a small infrared target (surrounded by the red rectangle) has positive contrast in all directions,  while a high-intensity edge (surrounded by the  yellow rectangle) does not have  such  property. Hence, using the minimum  value of the directional AAGD algorithm will effectively enhance false-alarm suppression ability. The following procedure describes the saliency map construction using the proposed method.

\begin{figure}[!t]
\centering
\includegraphics[width=0.3\textwidth]{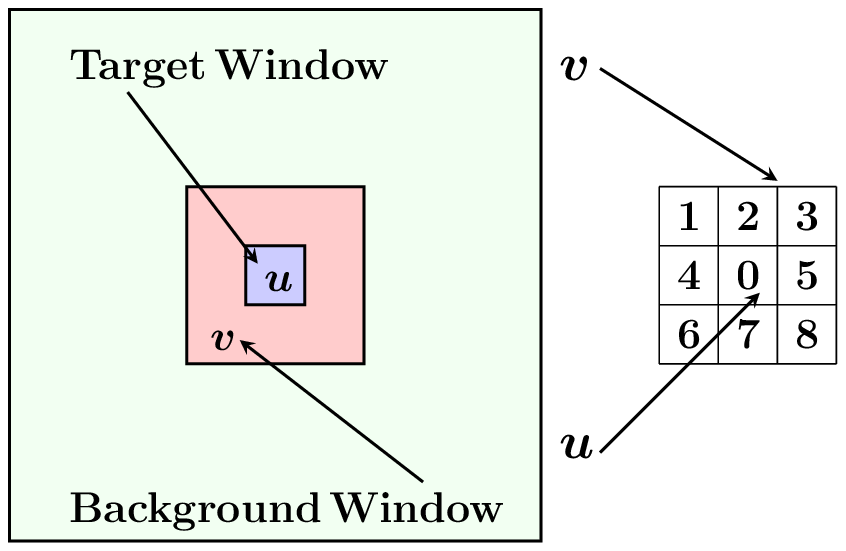}
\caption{The target and background window construction in the proposed method}
\label{fig:windows}
\end{figure}
\begin{figure*}[!t]
	\centering
	\subfloat[]{\includegraphics[width=1.8in]{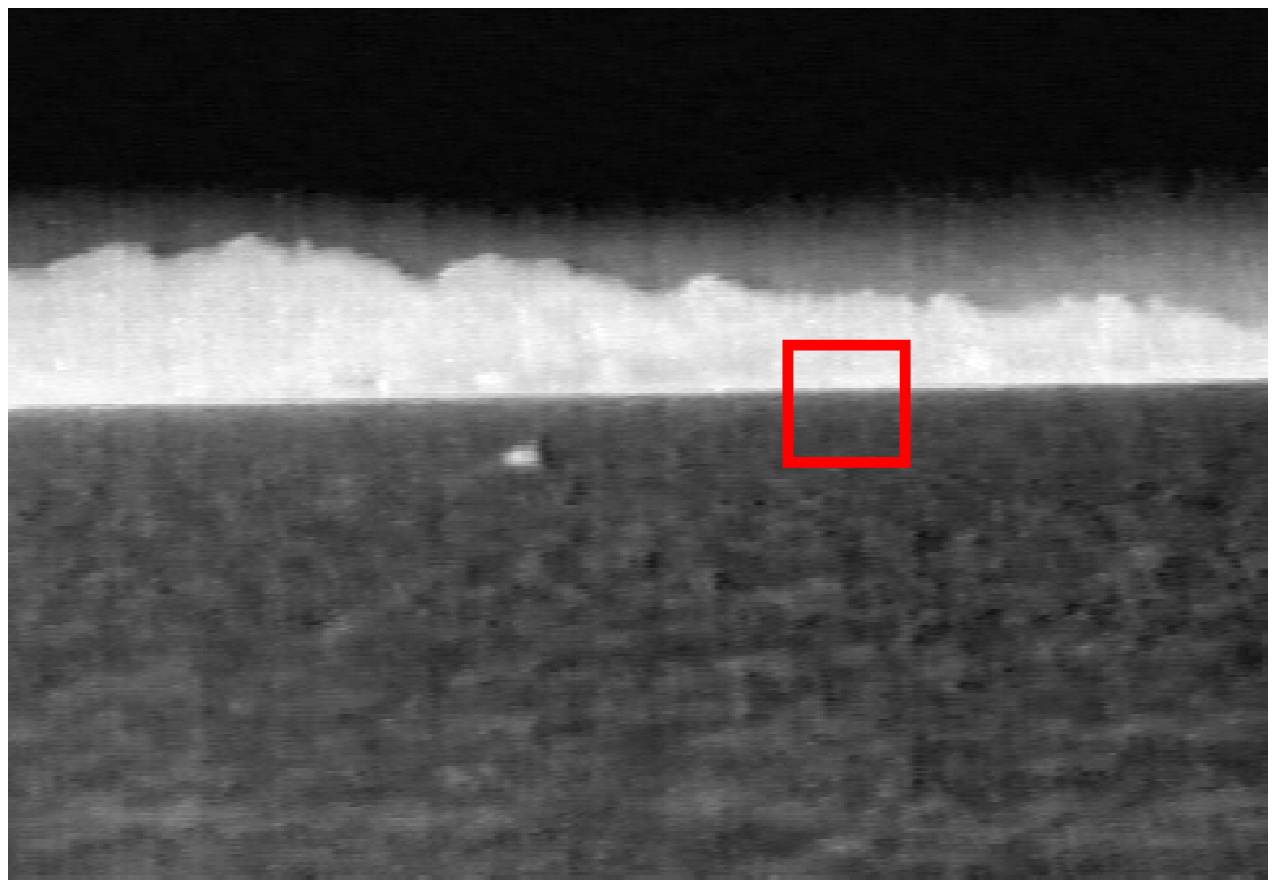}%
		\label{fig:edge_suppression_orig}}
	~~~~~~~~~~
	\subfloat[]{\includegraphics[width=1.8in]{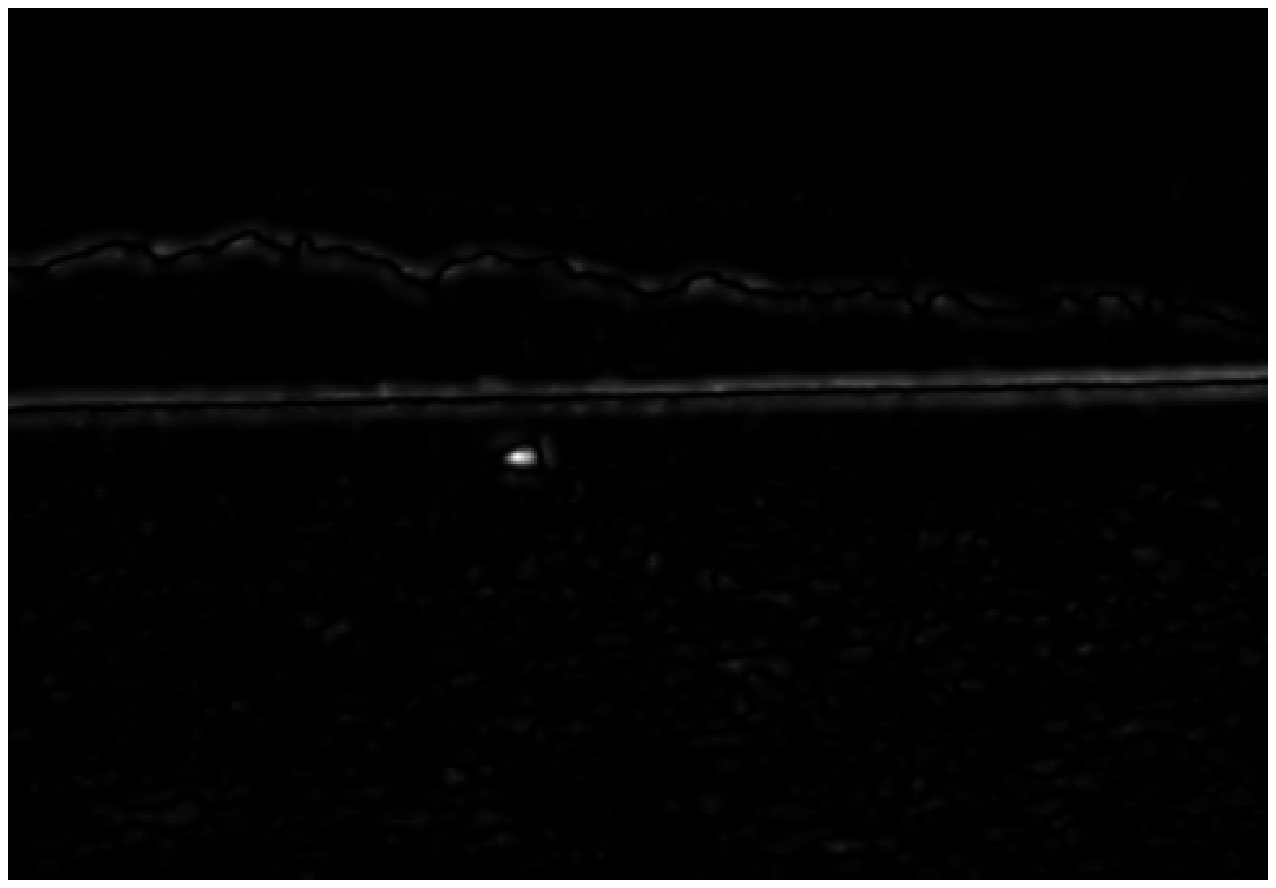}%
		\label{fig:edge_suppression_aagd}}
	~~~~~~~~~~
	\subfloat[]{\includegraphics[width=1.8in]{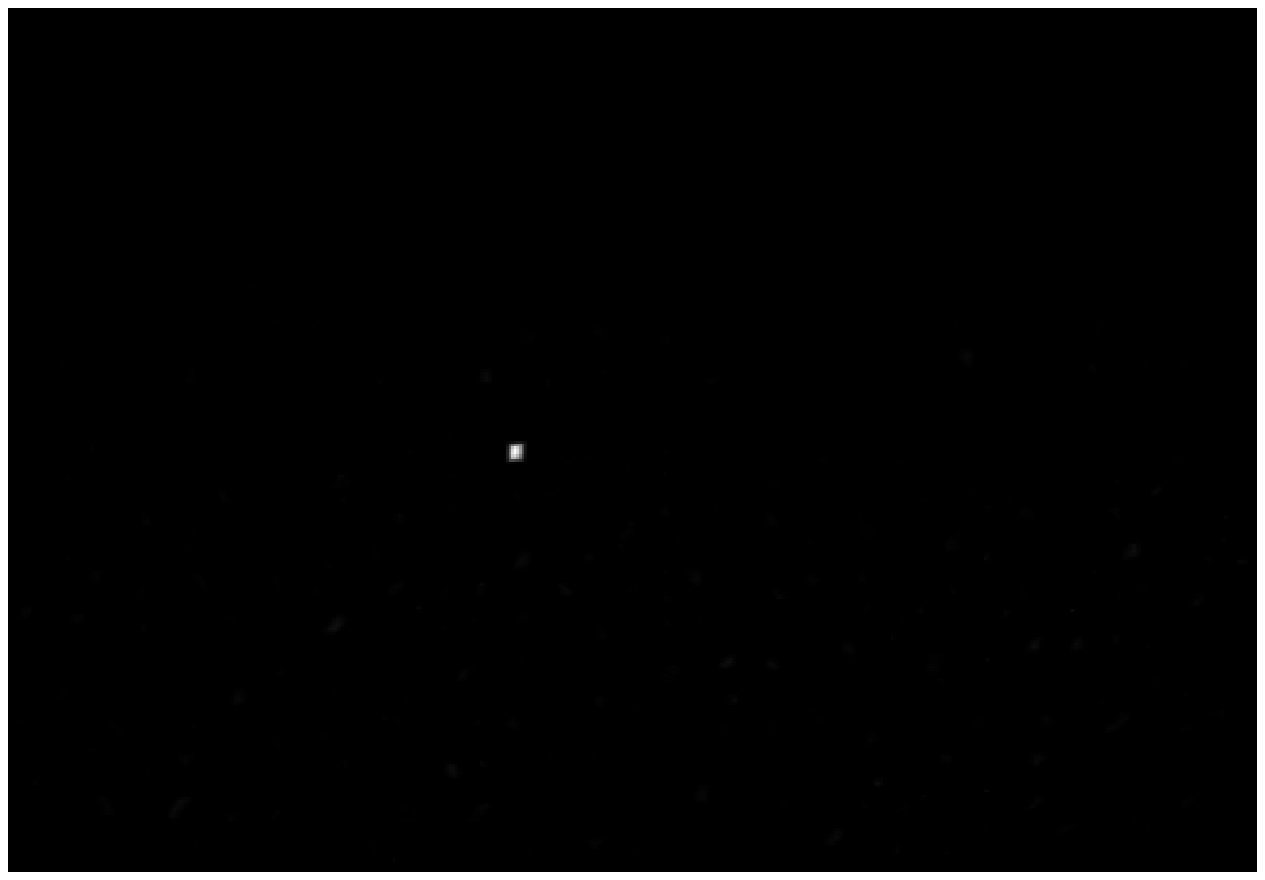}%
		\label{fig:edge_suppression_prop}}
	
	\subfloat[]{\includegraphics[width=2.2in]{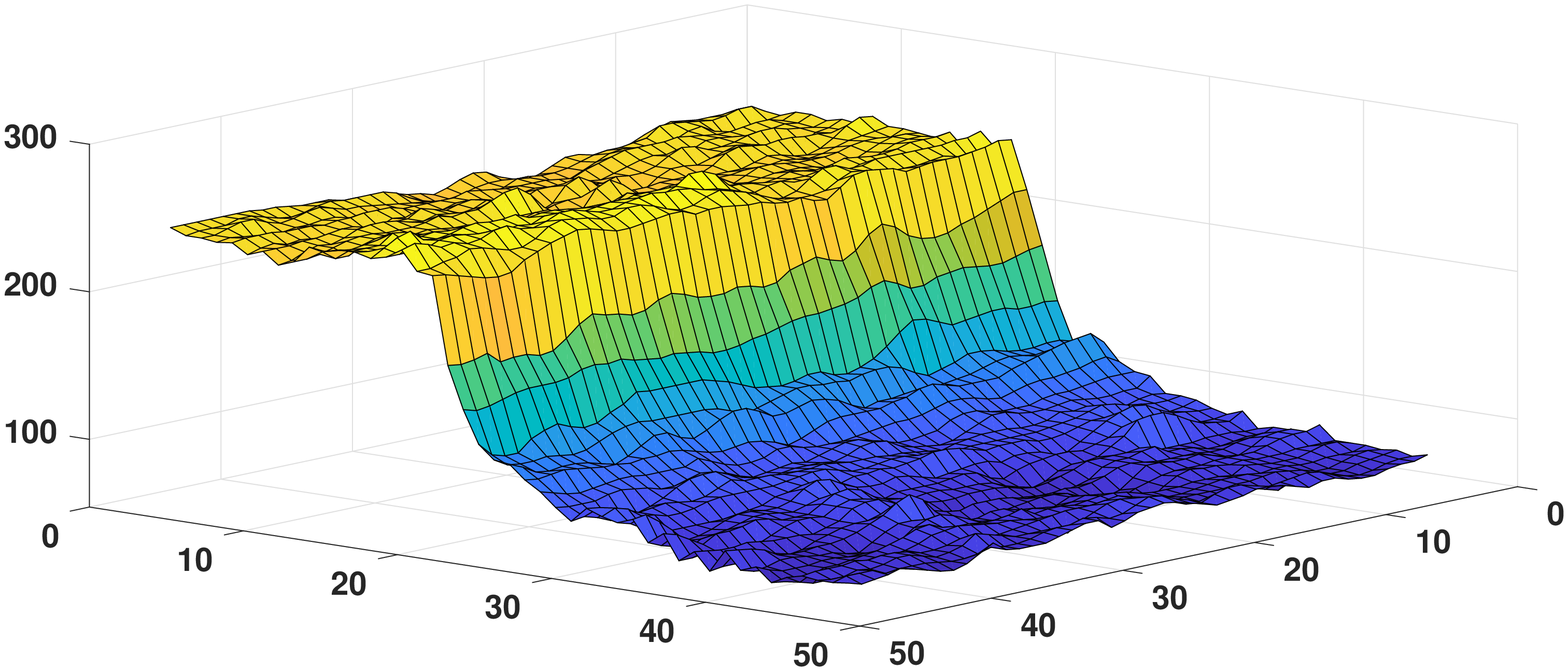}%
	\label{fig:edge_response_orig_3d}}
	~
	\subfloat[]{\includegraphics[width=2.2in]{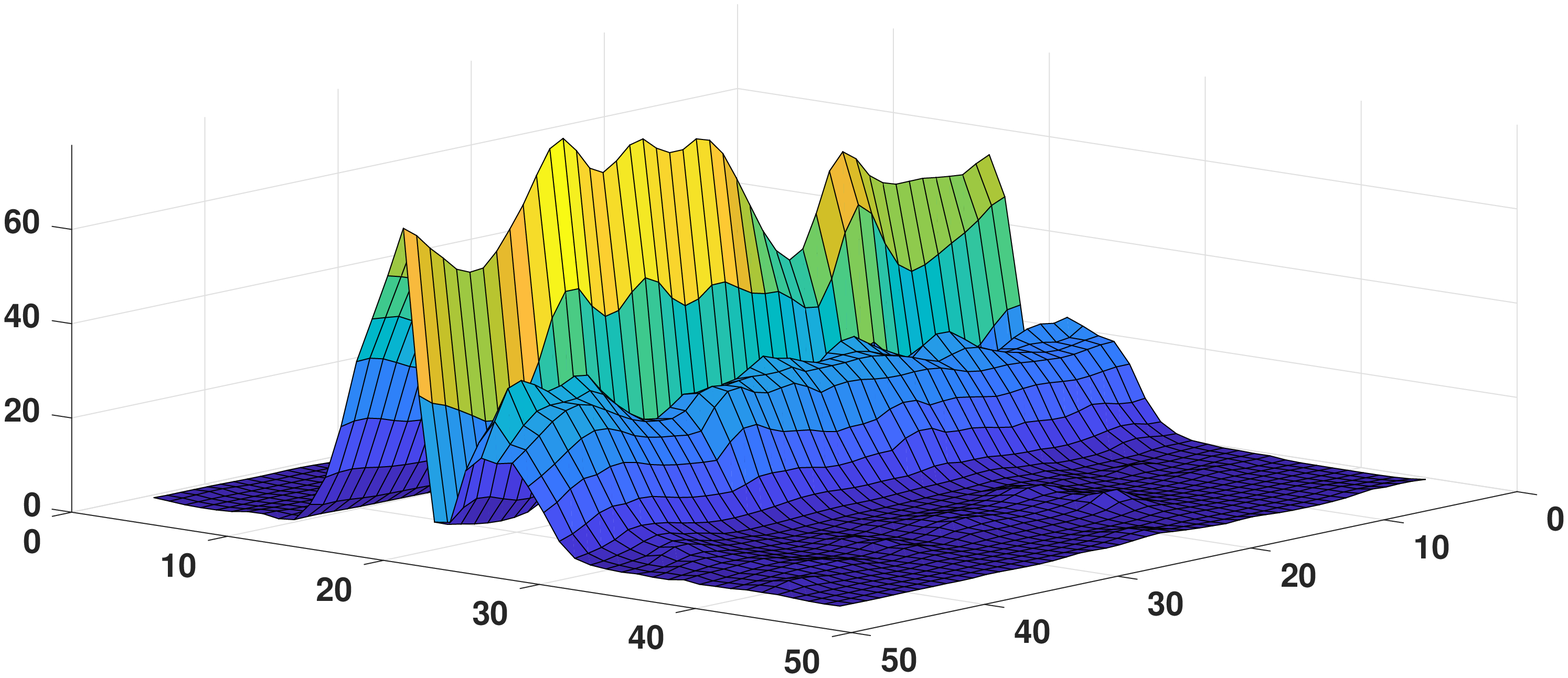}%
		\label{fig:edge_response_aagd_3d}}
	~
	\subfloat[]{\includegraphics[width=2.2in]{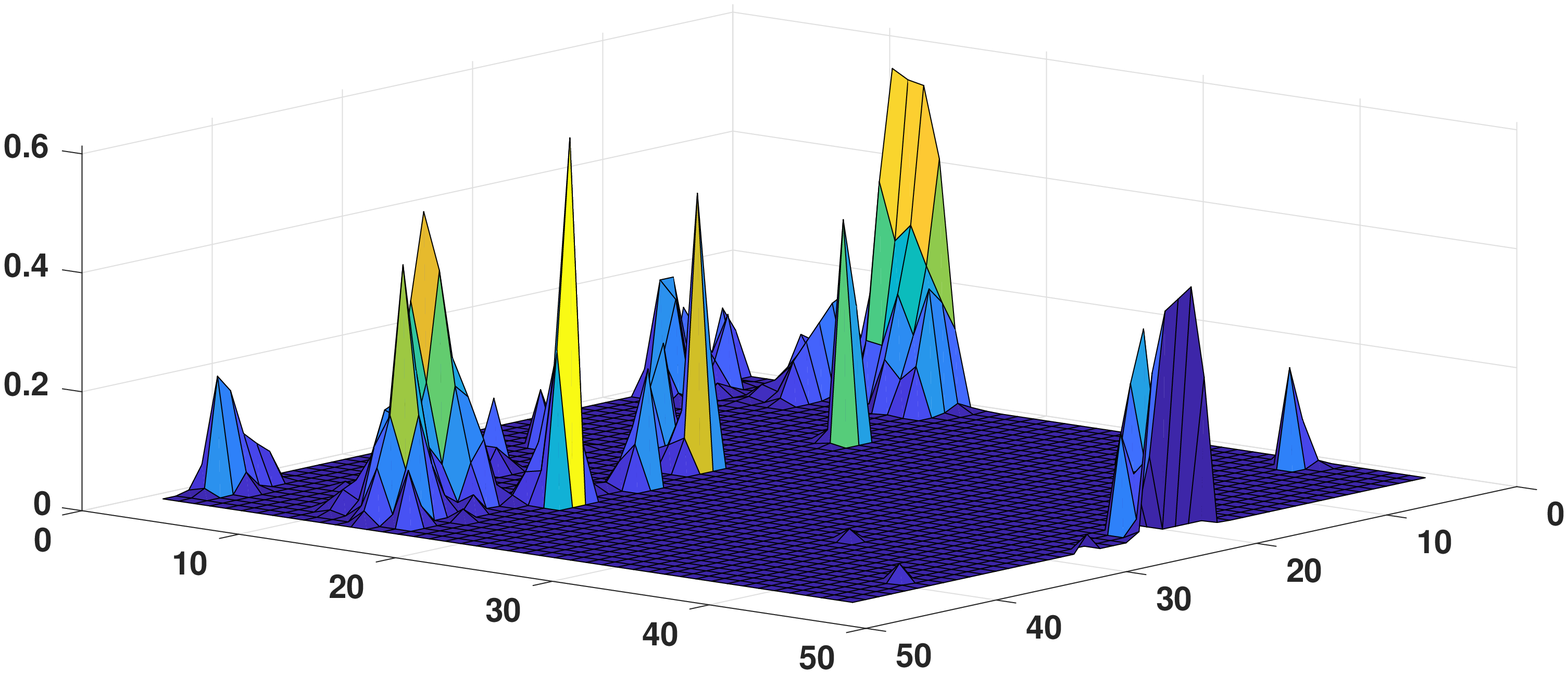}%
		\label{fig:edge_response_prop_3d}}
	
	\caption{The output response of the AAGD as well as the proposed algorithm to a sharp edge: {\bf a)} original input image, {\bf b)} the filtering result of the AAGD algorithm,  {\bf c)} the filtering result of the proposed algorithm, {\bf d)} 3-D surface of the red rectangle in the input image, {\bf e)} 3-D surface of the red rectangle in the filtering result of the AAGD algorithm,  {\bf f)} 3-D surface of the red rectangle in the filtering result of the proposed algorithm. Note that, the gray intensity of all images are normalized to $[0-255]$ interval.}
	\label{fig:edge_response}
\end{figure*}
As depicted in \autoref{fig:windows}, two nested windows are assigned for each pixel $(i,j)$ in the input image. The internal one, $u$, denotes the target window and its size is related to the small infrared target size (typically smaller than $9\times9$ based on SPIE definition \cite{zhang2019infrared}). The external window, $v$, stands for local background window and its size is $3$ times of the size of the target window. As shown in the figure, the external window is divided to $9$ equal cells (obviously the $cell_0$ is the target window $u$). The average intensity of the pixels of the cell $k$  is denoted by $m_k$ and calculated as follows:
\begin{equation}
m_k(i,j)=\frac{1}{N_{cell}}\sum_{(s,t)\in cell_k}I^k(s,t) \quad\quad i=0,1,\ldots,8
\end{equation}  
where, $I^k(s,t)$ is the gray value of the pixel $(s,t)$ in $k^{th}$ cell when the center of $cell_0$  lies on pixel $(i,j)$. Also, $N_{cell}$ is the number of pixels in each cell. 
The AAGD algorithm for eight different directions can be computed as: 
\begin{equation}
AAGD_k(i,j)=\left( m_0(i,j)-m_k(i,j) \right)^2 \quad\quad k=1,\ldots,8
\end{equation}
In order to address the hole-like objects issue in the AAGD algorithm and reduce false alarms, simply the resulting negative contrast are suppressed  using the following thresholding procedure:
\begin{equation}
D_k(i,j)=AAGD_k(i,j)\times H\left(m_0(i,j)-m_k(i,j)\right)  \quad k=1,\ldots,8
\end{equation}
where $H(\cdot)$ is the Heaviside step function and expressed as:
  \begin{equation}
  H(x) = 
  \begin{cases} 
  1 & \quad \quad \quad  x \geq 0 \\
  0 & \quad \quad \quad  x < 0
  \end{cases}.
  \end{equation}
  
  As mentioned before, unlike the structural background clutters,  small targets have positive contrast in all directions. Hence, considering minimum directional response as the final output will enhance clutter rejection capability, while the target detection ability remains unchanged. Therefore, in this paper, absolute directional  mean difference (ADMD) value is defined as:
\begin{equation}
ADMD(i,j)=\min\left\{D_1(i,j),D_2(i,j),\ldots,D_8(i,j)\right\} 
\end{equation}  
  
This procedure can be done through several scales (different cell size) for multi-scale processing. The output of the multi-scale ADMD algorithm can be expressed as:  
\begin{equation}
\begin{split}
& MS\_ADMD(i,j)=\\
&\max\left\{ADMD^1(i,j),ADMD^2(i,j),\ldots,ADMD^S(i,j)\right\} 
\end{split}
\end{equation}  
where $ADMD^s$ denotes the ADMD output at the $s^{th}$ scale.

\subsection{Clutter rejection mechanism}
\label{sec:crm}
\autoref{fig:edge_response} shows the output response of the AAGD as well as the proposed algorithm to an infrared scene including a sharp cloud edge, which the gray intensity of all images are normalized to $[0-255]$ interval. As shown in the figure, the proposed algorithm effectively eliminates the high-intensity edge. The  gray intensity of the red rectangle does not exceed 0.6 (\autoref{fig:edge_response_prop_3d}). However, the AAGD algorithm has a relatively strong response in that area which is demonstrated as two parallel lines in the figure. As shown in \autoref{fig:edge_response_aagd_3d}, the maximum gray intensity of the area related to the red rectangle is about 60 which is approximately 100 times greater than the ADMD output.

\begin{figure}[!t]
	\centering
	\subfloat[]{\includegraphics[width=1.5in]{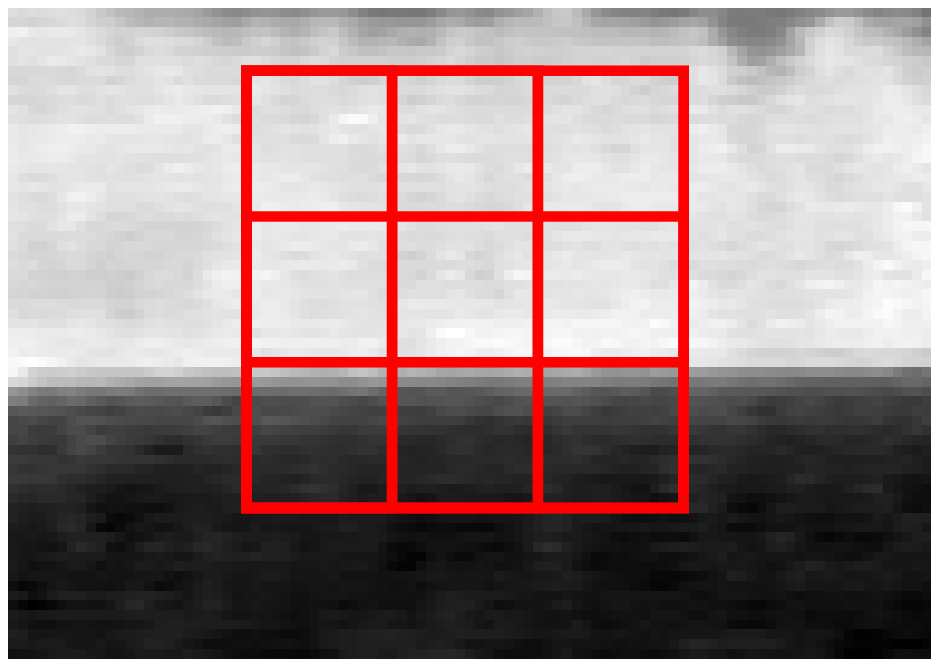}%
		\label{fig:aagd_edge_positive}}
	~~~~~~~~~~~
	\subfloat[]{\includegraphics[width=1.5in]{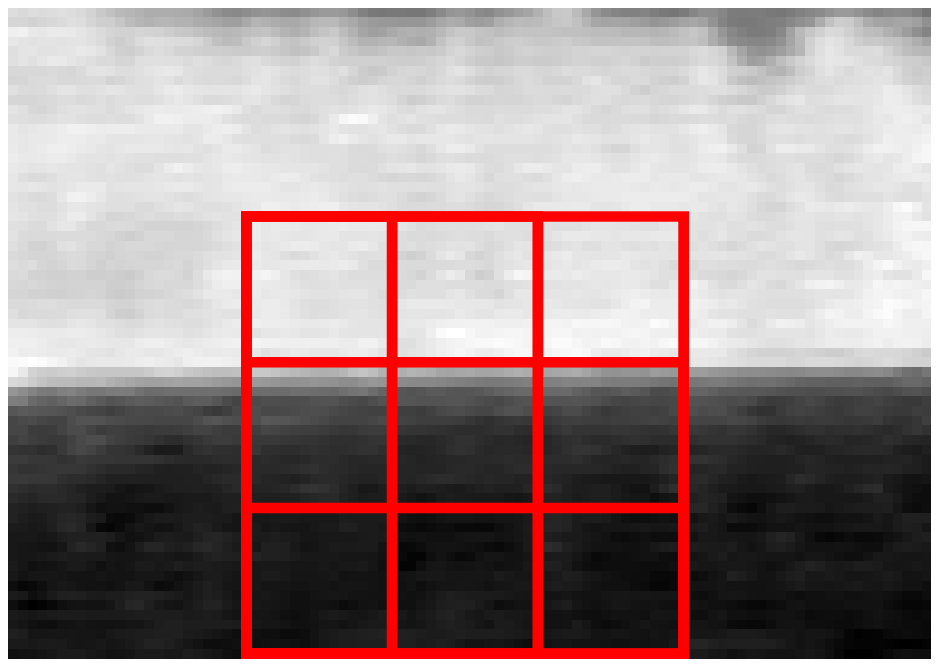}%
		\label{fig:aagd_edge_negative}}

	\caption{Two different scenarios to explain AAGD deficiency: {\bf a)} the target window has positive contrast, {\bf b)} the target window has negative contrast.}
	\label{fig:edge_response_analysis}
\end{figure}

In order to investigate the behavior of both AAGD and ADMD algorithms facing sharp edges, a brief analysis is provided here (\autoref{fig:edge_response_analysis}). Without loss of generality, let assume that the  area under test  just has two gray levels ($g_b$ and $g_d$ values for bright and dark areas, respectively). Also, let assume that both target and background windows are identical for  both algorithms. As the final assumption, let assume that the algorithms are constructed just in a single scale. In the first scenario (\autoref{fig:aagd_edge_positive}), the  target window completely lies on bright area. In this situation, the output of the AAGD algorithm can be calculated as:
\begin{equation}
\begin{split}
out_{AAGD}&=\left(g_b-\left(\frac{5\times g_b+3\times g_d}{8}\right)\right)^2 \\
&=\left(\frac{3}{8}\left(g_b-g_d \right)\right)^2=\frac{9}{64}\Delta g^2
\end{split}
\label{eq:aagd_positive_out}
\end{equation} 

According to \autoref{eq:aagd_positive_out}, as $\Delta g$ increases (the edge becomes more sharper), the output of the AAGD algorithm becomes more intensified (\autoref{fig:edge_response_aagd_3d}).

Considering \autoref{fig:aagd_edge_positive} as the input scenario, there are only two types of directional output in ADMD algorithm. $D_b$  for the directions along with bright background cells and $D_d$ for the directions along with dark background cells. These two outputs are calculated as follows:
\begin{equation}
\begin{split}
&D_b=H\left(g_b-g_b\right)\times\left(g_b-g_b\right)^2=0\\
&D_d=H\left(g_b-g_d\right)\times\left(g_b-g_d\right)^2=\Delta g^2
\end{split}
\end{equation}

Therefore,
\begin{equation}
out_{ADMD}=\min\left\{D_b,D_d\right\}=\min\left\{0,\Delta g^2\right\}=0
\label{eq:prop_positive_out}
\end{equation}

When the target window has negative contrast  compared to the bright background cells (\autoref{fig:aagd_edge_negative}), the output of AAGD algorithm is computed as follows:
\begin{equation}
\begin{split}
out_{AAGD}&=\left(g_d - \left(\frac{3\times g_b+5\times g_d}{8}\right)\right)^2\\
&=\left(\frac{3}{8}\left(g_d-g_b \right)\right)^2=\frac{9}{64}\Delta g^2
\end{split}
\label{eq:aagd_negaive_out}
\end{equation}

\autoref{eq:aagd_negaive_out} demonstrates that, similar to the former scenario (\autoref{fig:aagd_edge_positive}), the AAGD algorithm returns a strong response for the non-target area. When the center of the target window exactly lies on the edge, the AAGD output takes zero value. This is why there is a dark area between two parallel lines in the AAGD output (\autoref{fig:edge_suppression_aagd}).

The output of the ADMD algorithm to \autoref{fig:aagd_edge_negative} scenario is expressed as:
\begin{equation}
\begin{split}
&D_b=H\left(g_d-g_b\right)\times\left(g_d-g_b\right)^2=0\\
&D_d=H\left(g_d-g_d\right)\times\left(g_d-g_d\right)^2=0
\end{split}
\end{equation}

Therefore,
\begin{equation}
out_{ADMD}=\min\left\{D_b,D_d\right\}=\min\left\{0,0\right\}=0
\label{eq:prop_negaive_out}
\end{equation}

According to   \autoref{eq:prop_positive_out} and \autoref{eq:prop_negaive_out}, one can conclude that the proposed algorithm effectively eliminates high-intensity edges. It is worth  mentioning that since the small target has positive contrast in all directions, minimum selection strategy does not degrade detection ability.
\subsection{Noise suppression ability}
\begin{figure}[!t]
	\centering
	\subfloat[]{\includegraphics[width=3in]{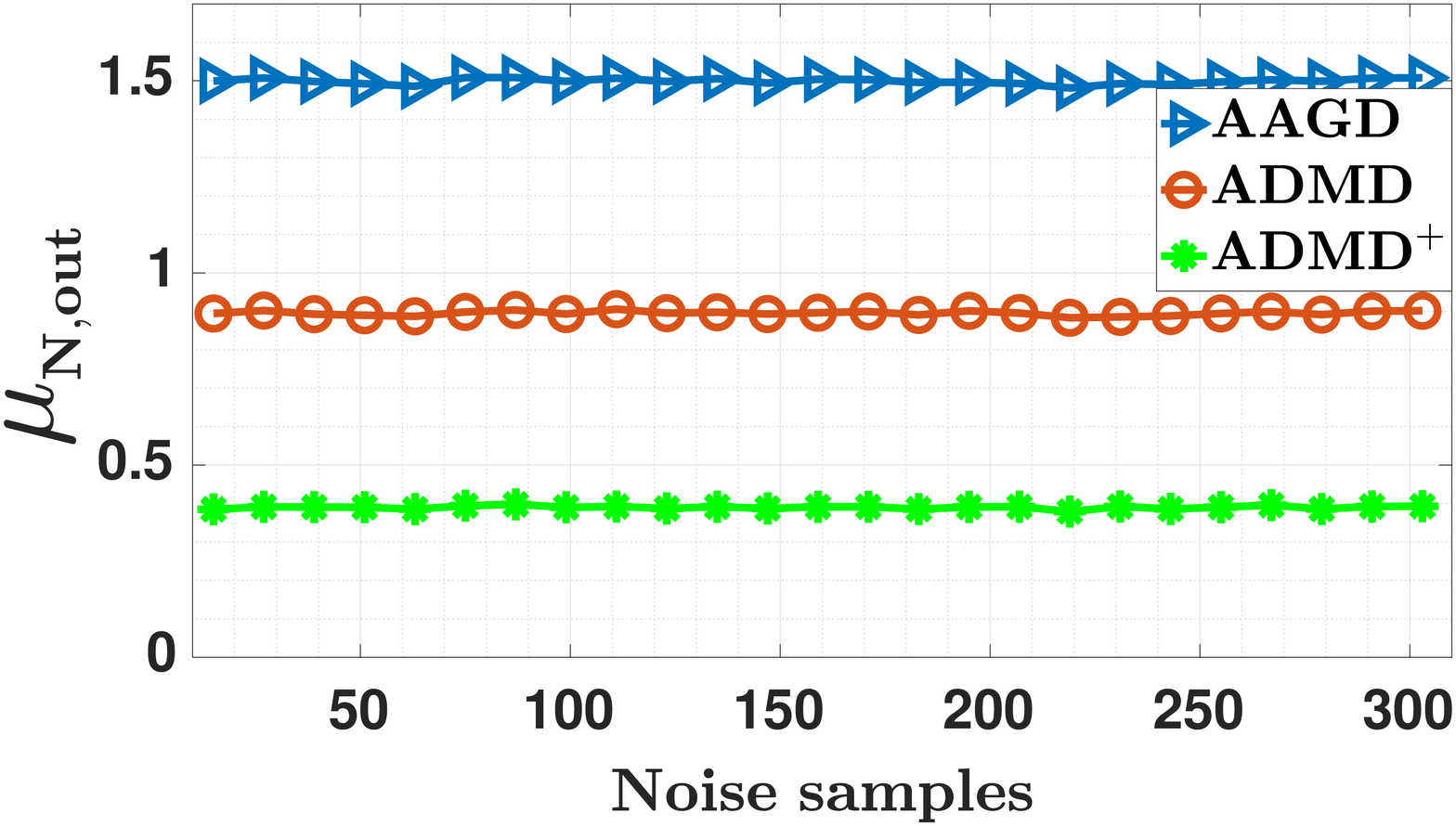}%
		\label{fig:noise_mu}}
	~~~~~~~~~~~
	\subfloat[]{\includegraphics[width=3in]{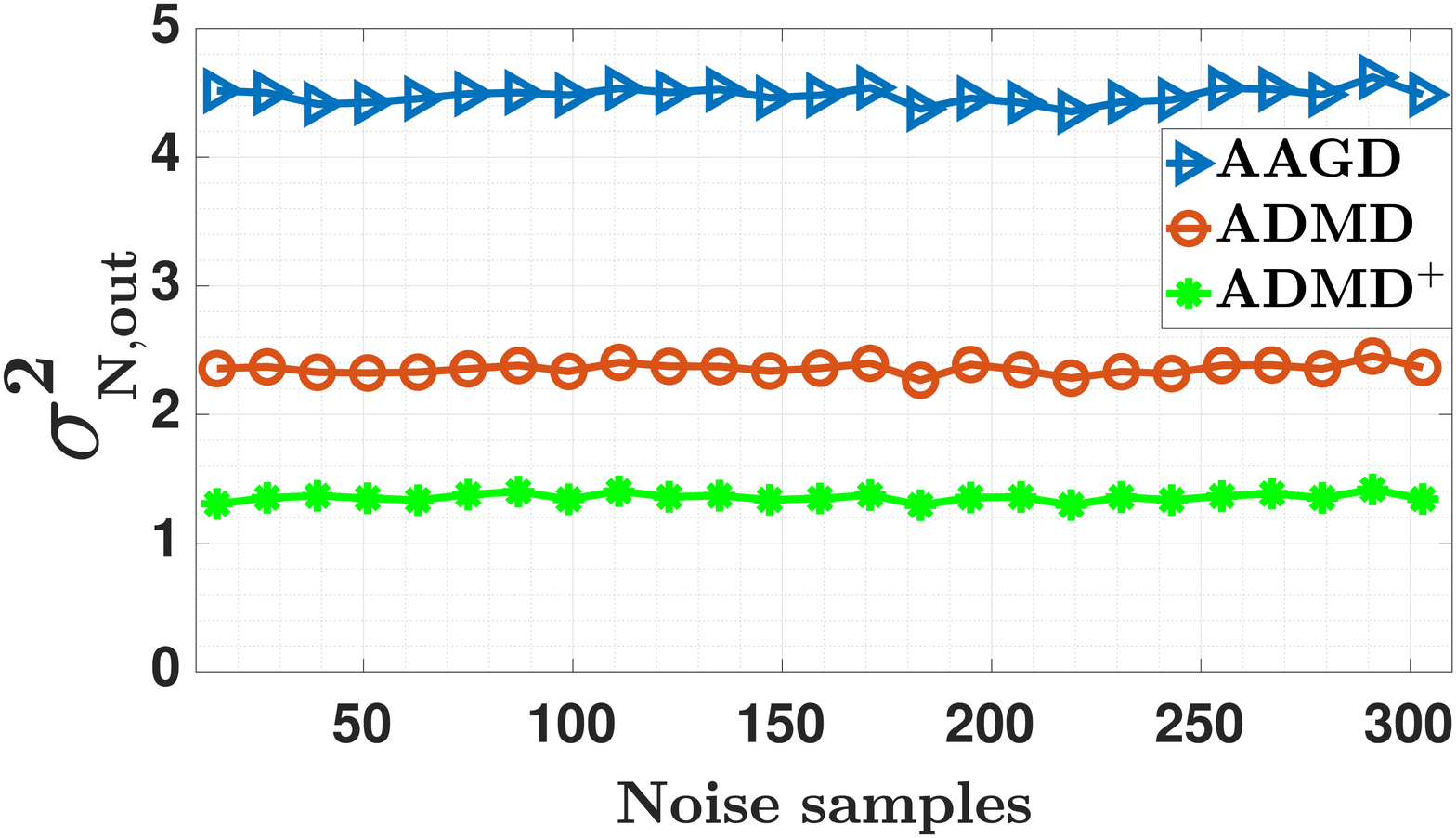}%
		\label{fig:noise_var}}

	\caption{Output noise response for different algorithms (1-D cases): {\bf a)} the mean value of output noise response, {\bf b)} the variance of output noise response. $\text{ADMD}$ and $\text{ADMD}^+$ denote the proposed algorithm without and with negative value elimination, respectively. The cell size (target window) and the size of AAGD background window are set to $9\times 9$ and $27\times 27$. Also, the standard deviation of the input noise ($\sigma_{_{N,in}}$) is set to $3$. }
	\label{fig:noise_response_analysis}
\end{figure}

At first sight, because of the smaller local background cell,   it may seem that the proposed method has lower background noise rejection ability compared to the AAGD algorithm. However, the both minimum selection operator and negative value elimination strategy in the ADMD algorithm significantly improve the  noise suppression ability of the proposed method. To compare the noise suppression ability of the AAGD as well as the proposed algorithm,  1-D additive white Gaussian noise (AWGN) is used as an input signal. The simulation  is carried out 100000 times for the input noise signal and the mean and variance values of the output are taken into account.
The mean and variance of the output noise response are shown in \autoref{fig:noise_response_analysis}. As shown in  the figure, the proposed algorithm has better performance in background noise suppression. 

\begin{figure}[!t]
	\centering
	\subfloat[]{\includegraphics[width=3in]{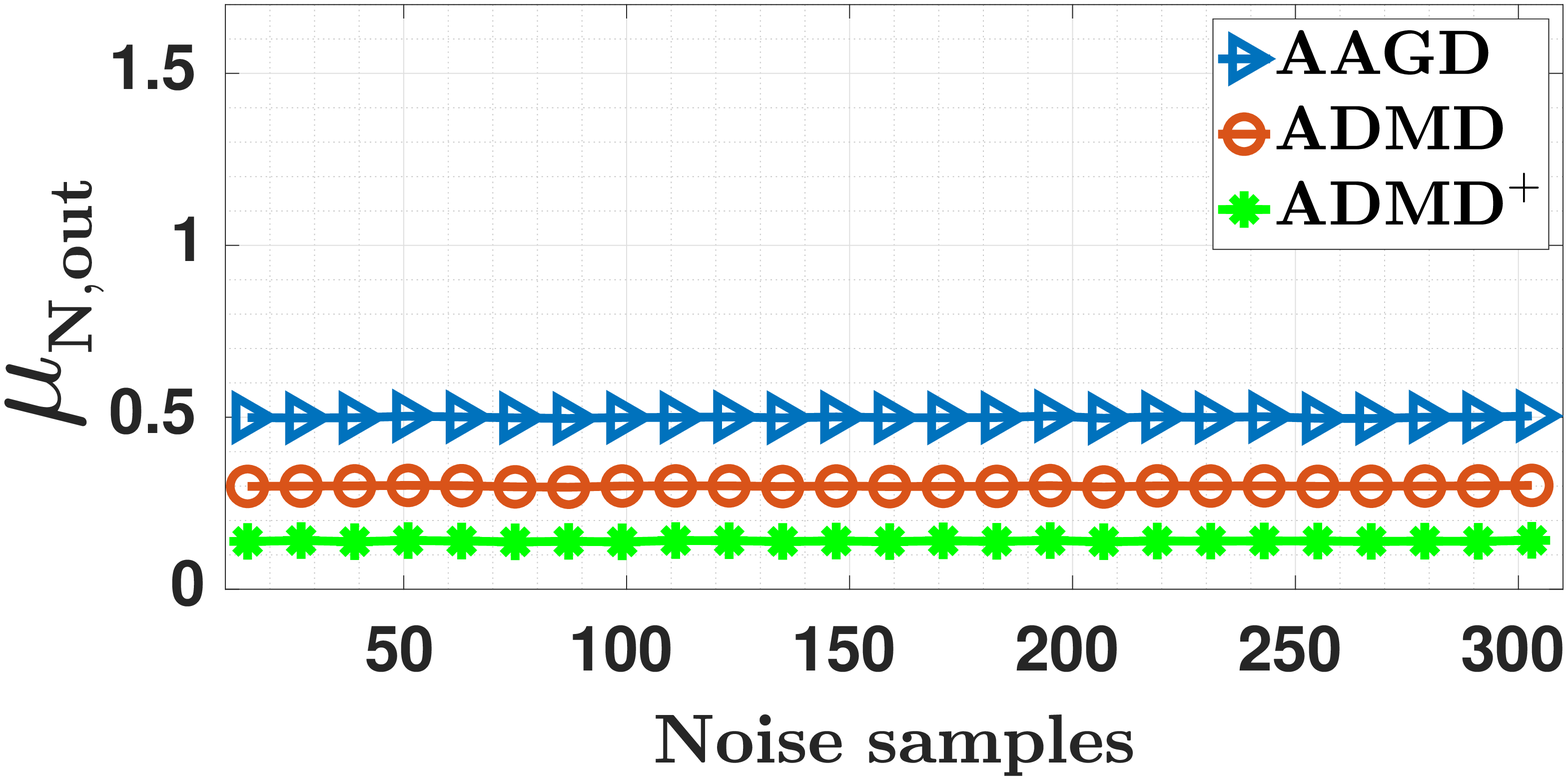}%
		\label{fig:noise_mu3}}
	~~~~~~~~~~~
	\subfloat[]{\includegraphics[width=3in]{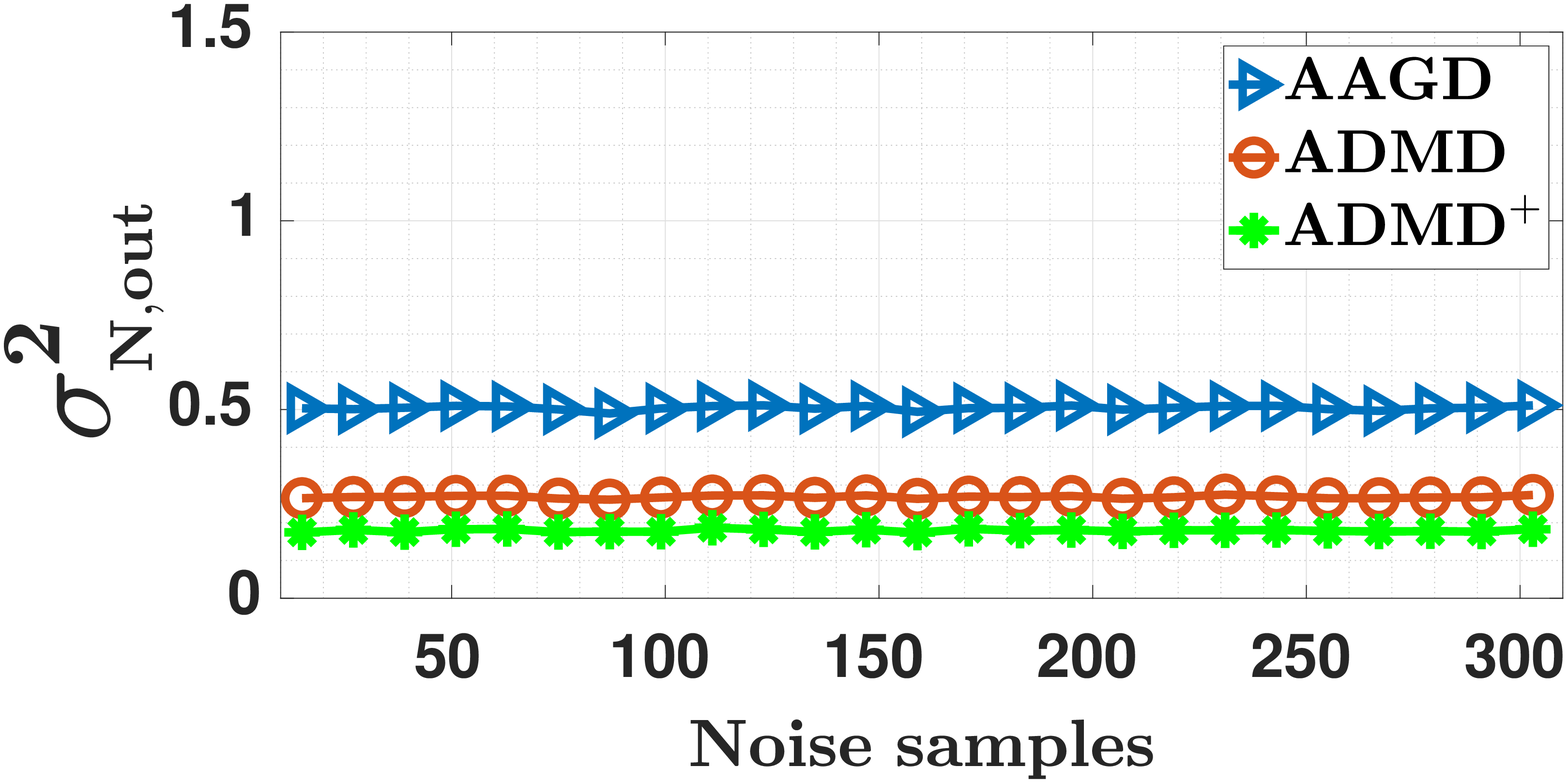}%
		\label{fig:noise_var3}}
\\
	\subfloat[]{\includegraphics[width=3in]{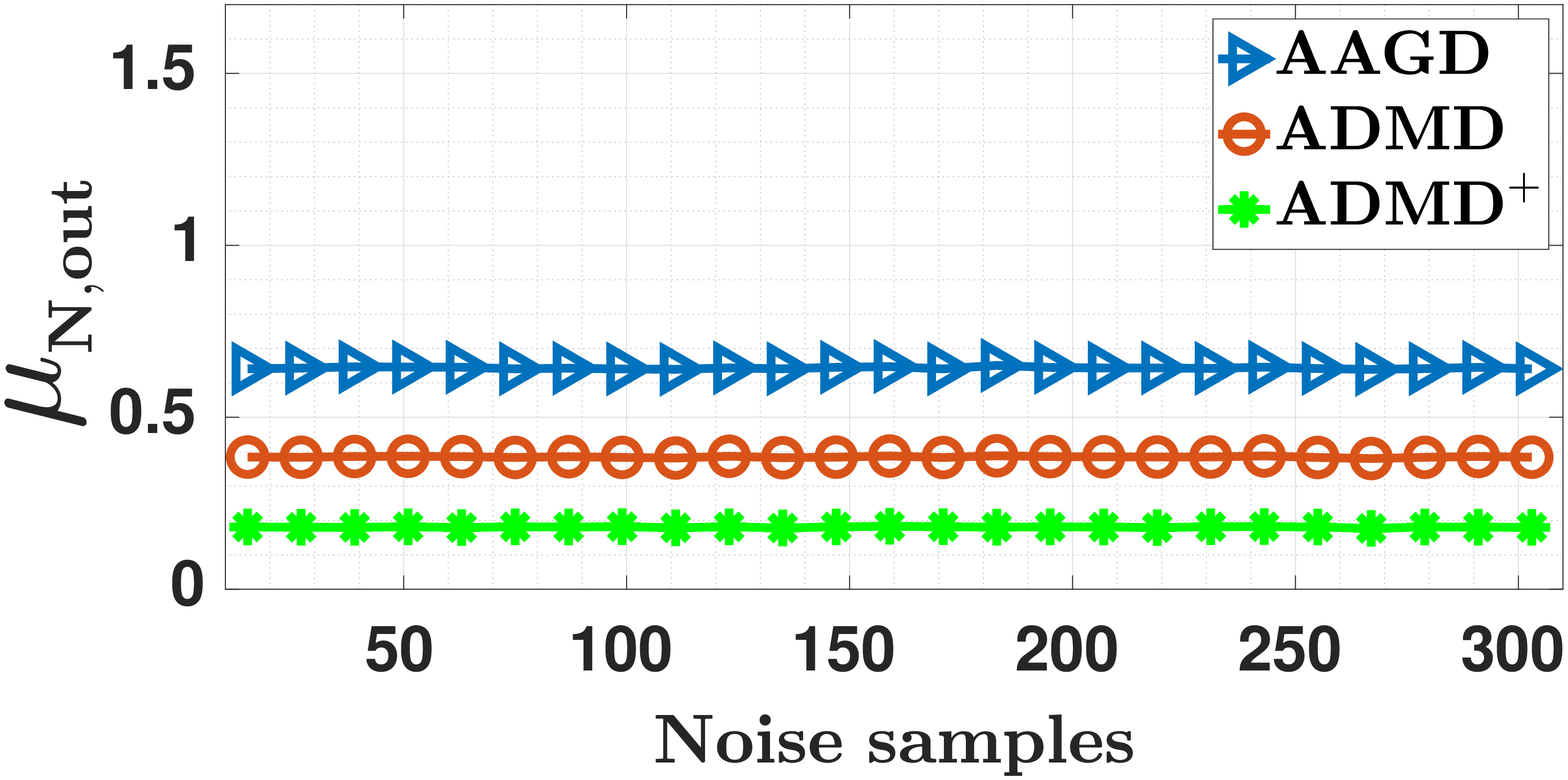}%
		\label{fig:noise_mu2}}
	~~~~~~~~~~~
	\subfloat[]{\includegraphics[width=3in]{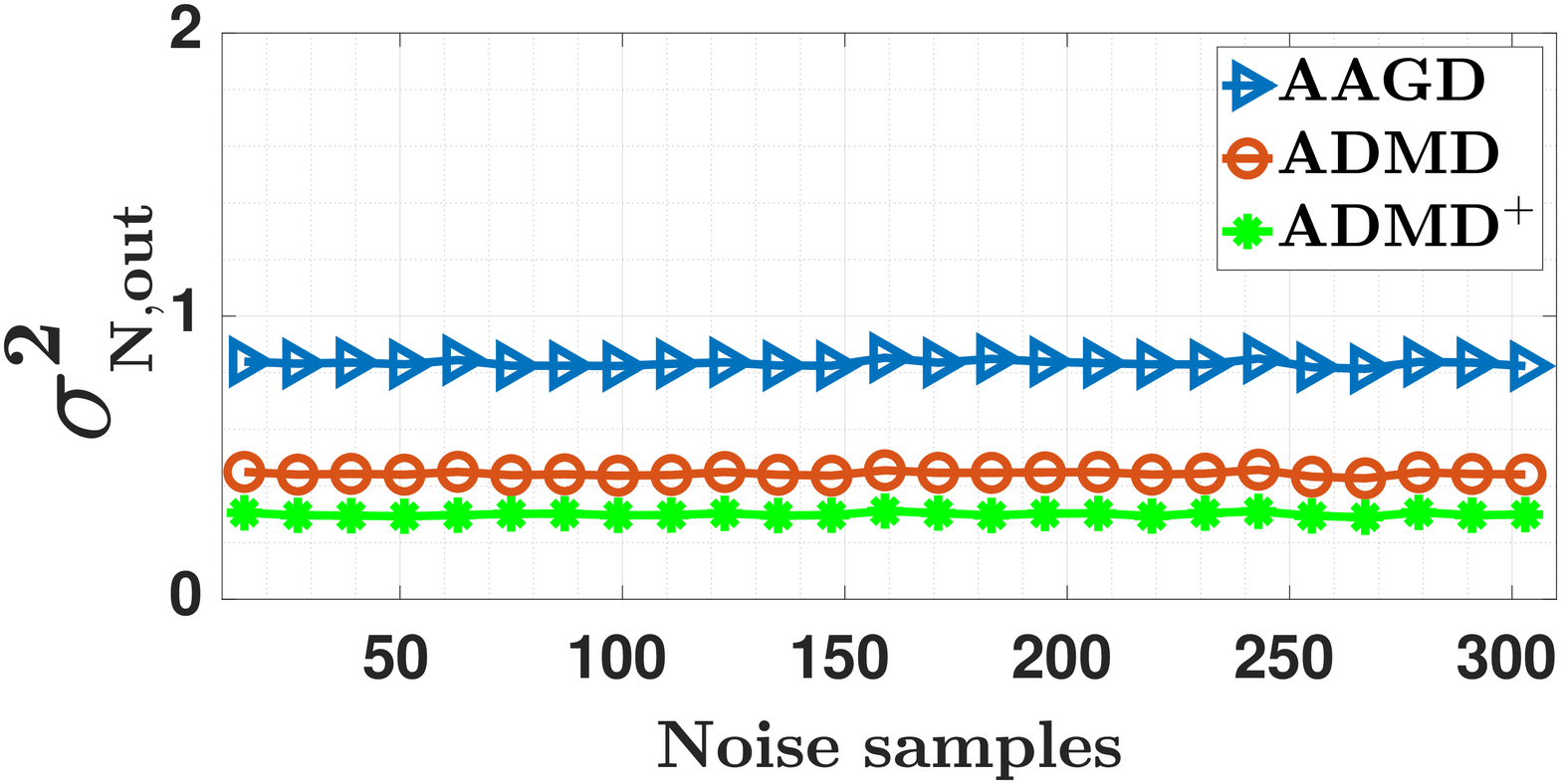}%
		\label{fig:noise_var2}}
	\caption{Output noise response for different algorithms (1-D cases): {\bf a)} the mean value of output noise response when the input noise has Poisson distribution, {\bf b)} the variance of output noise response when the input noise has Poisson distribution, {\bf c)} the mean value of output noise response when the input noise has Rayleigh distribution, {\bf d)} the variance of output noise response when the input noise has Rayleigh distribution. The both $\lambda$ (for Poisson distribution) and scale parameter (for Rayleigh distribution) are set to $3$. }
	\label{fig:noise_response_analysis2}
\end{figure}

In order to investigate how the proposed method copes with non-Gaussian noises, a similar simulation is carried out for Poisson as well as Rayleigh noises. As depicted in the \autoref{fig:noise_response_analysis2}, the proposed method outperforms the AAGD, regardless of the noise distribution. Note that, regardless of the distribution of the background noise, the detection algorithm should be able to eliminate the noise and enhance only the true small target area. Thus,  no prior assumption  is made about the distribution of the background noise. However,  modeling of background noise  distribution is necessary for image restoration and noise removal, where the visual quality of the infrared scene is important. The interested reader may be referred to \cite{gong2014image,stojanovic2016identification,prsicrobust} for further information.

\subsection{Efficient implementation}
In order to construct a single-scale ADMD algorithm, eight directional AAGD algorithms should be computed, separately. Performing this amount of operations needs high processing resources. In this section, the proposed algorithm is reformulated to reduce computational complexity.

Since structural background clutters do not have positive  contrast in all directions, considering the maximum average intensity of the background cells in contrast function formulation will mimic the behavior of the proposed algorithm. Therefore, maximum directional mean value $M_{dir}$ is defined as:
\begin{equation}
M_{dir}(i,j)=\max\left\{m_1(i,j),m_2(i,j),\ldots,m_8(i,j)\right\}
\end{equation}

Accordingly, absolute difference mean value for the pixel $(i,j)$ is calculated as follows:
\begin{equation}
ADM(i,j)=(m_0(i,j)-M_{dir}(i,j))^2
\end{equation}

By discarding negative values through thresholding operation, the final ADMD output will achieve: 
\begin{equation}
ADMD(i,j)=ADM(i,j)\times H\left(m_0(i,j)-M_{dir}(i,j)\right)
\end{equation}

\begin{figure}[!t]
\centering
\includegraphics[width=0.3\textwidth]{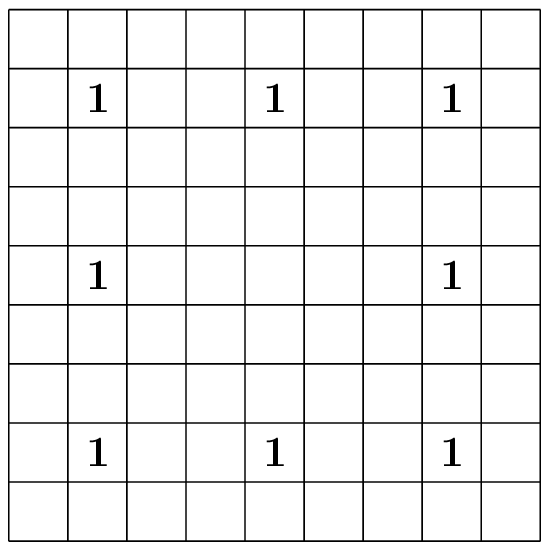}
\caption{A $9\times 9$ structural element to obtain maximum directional value for  $3\times 3$ target size (empty cells have zero values).}
\label{fig:SL}
\end{figure}

This new formulation can be efficiently implemented via local averaging and morphology dilation operators. To this end, the following procedure can be utilized to obtain the final saliency map:
\begin{enumerate}
\item The input infrared image is filtered using local averaging with proper neighborhood size ($3\times 3$, $5\times 5$, $7\times 7$, or $9\times 9$), $m_0$.
\item The filtered image $m_0$ is morphologically  dilated using proper structural element (e.g. \autoref{fig:SL}) to obtain maximum directional value, $M_{dir}$.
\item Then, the absolute difference mean is calculated, $ADM$.
\item Finally, the saliency map is constructed by eliminating negative contrasts, $ADMD$.
\end{enumerate}

\section{Simulation results}
\subsection{Detection ability analysis}
\begin{figure*}[h!]
	\centering
	\subfloat[]{\includegraphics[width=0.93in]{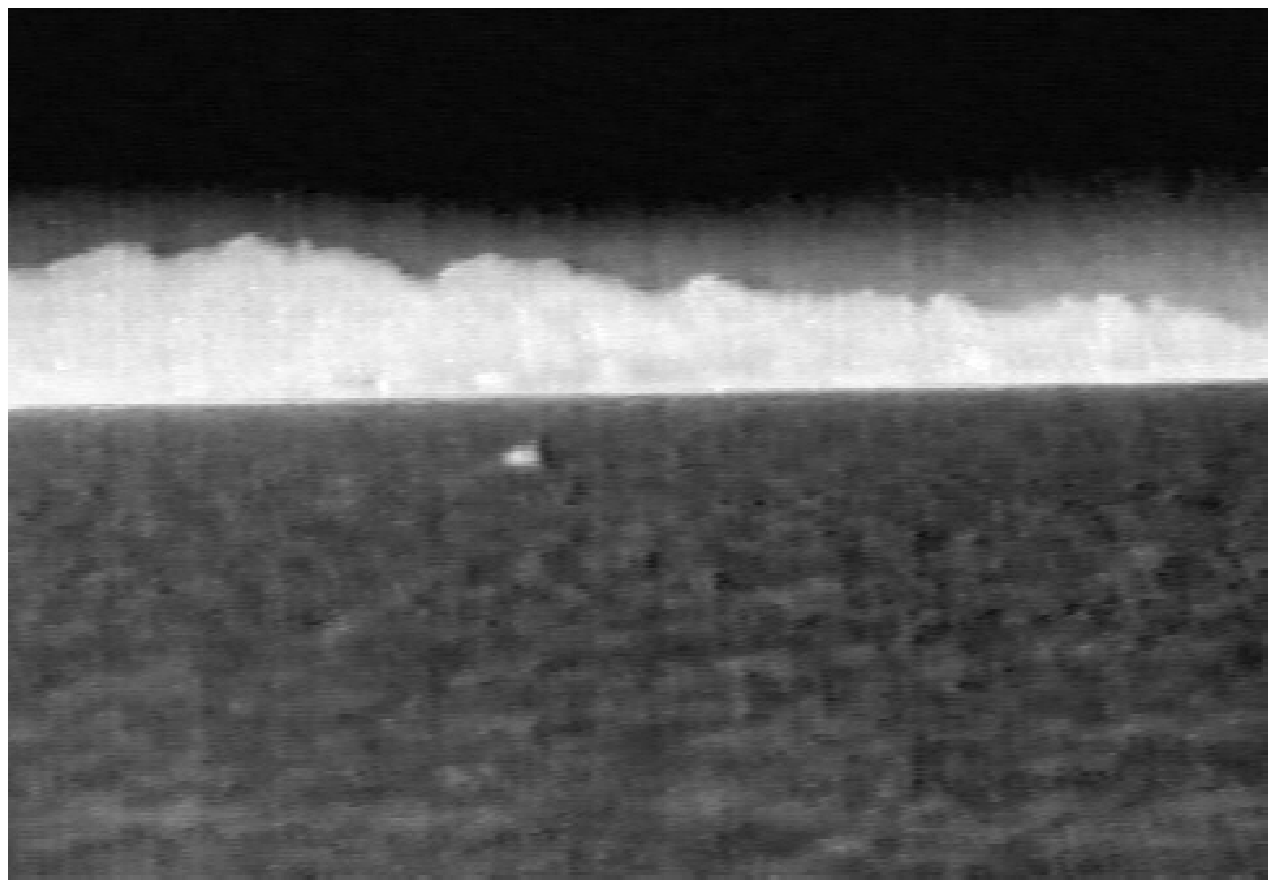}%
		\label{fig:original2}}
		~
		\subfloat[]{\includegraphics[width=0.93in]{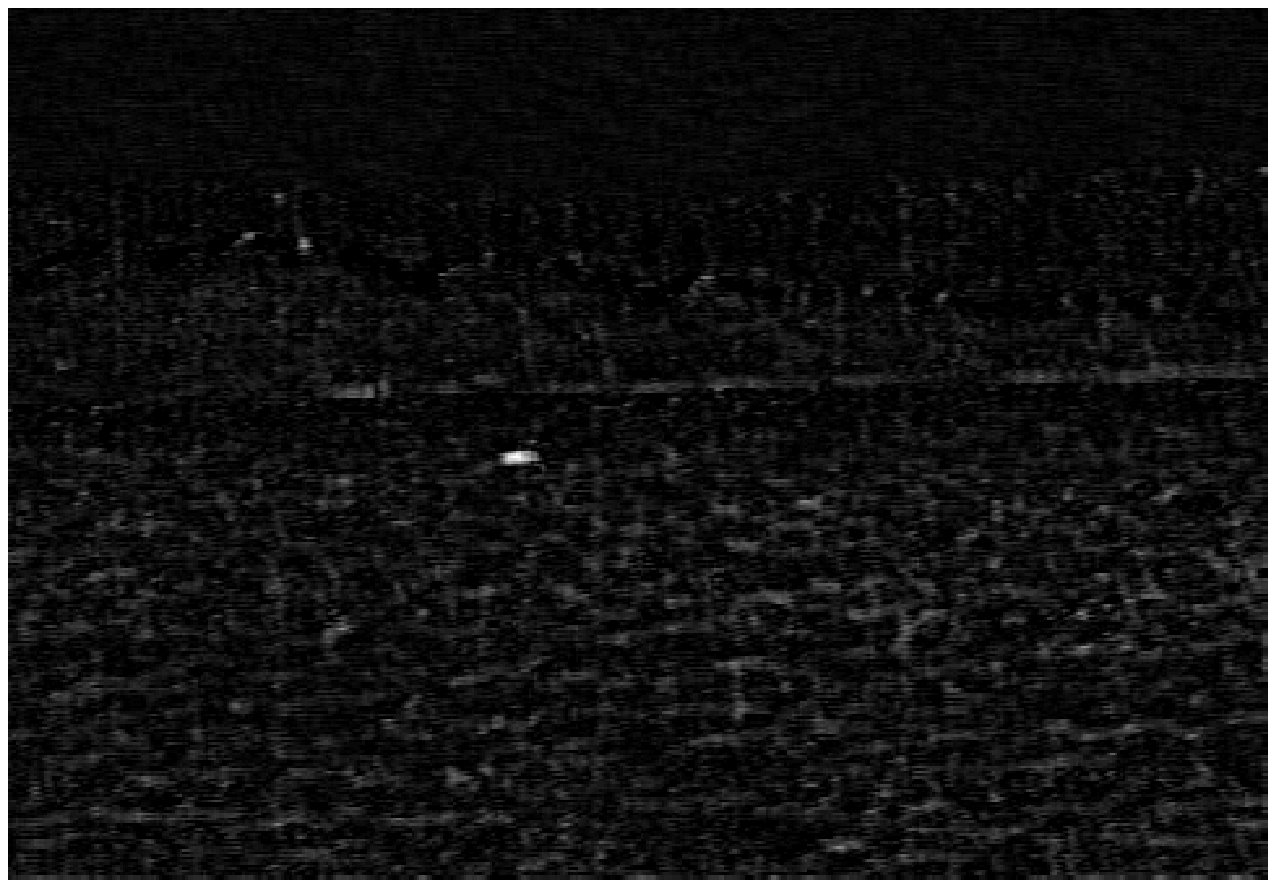}%
		\label{fig:top2}}
	~
	\subfloat[]{\includegraphics[width=0.93in]{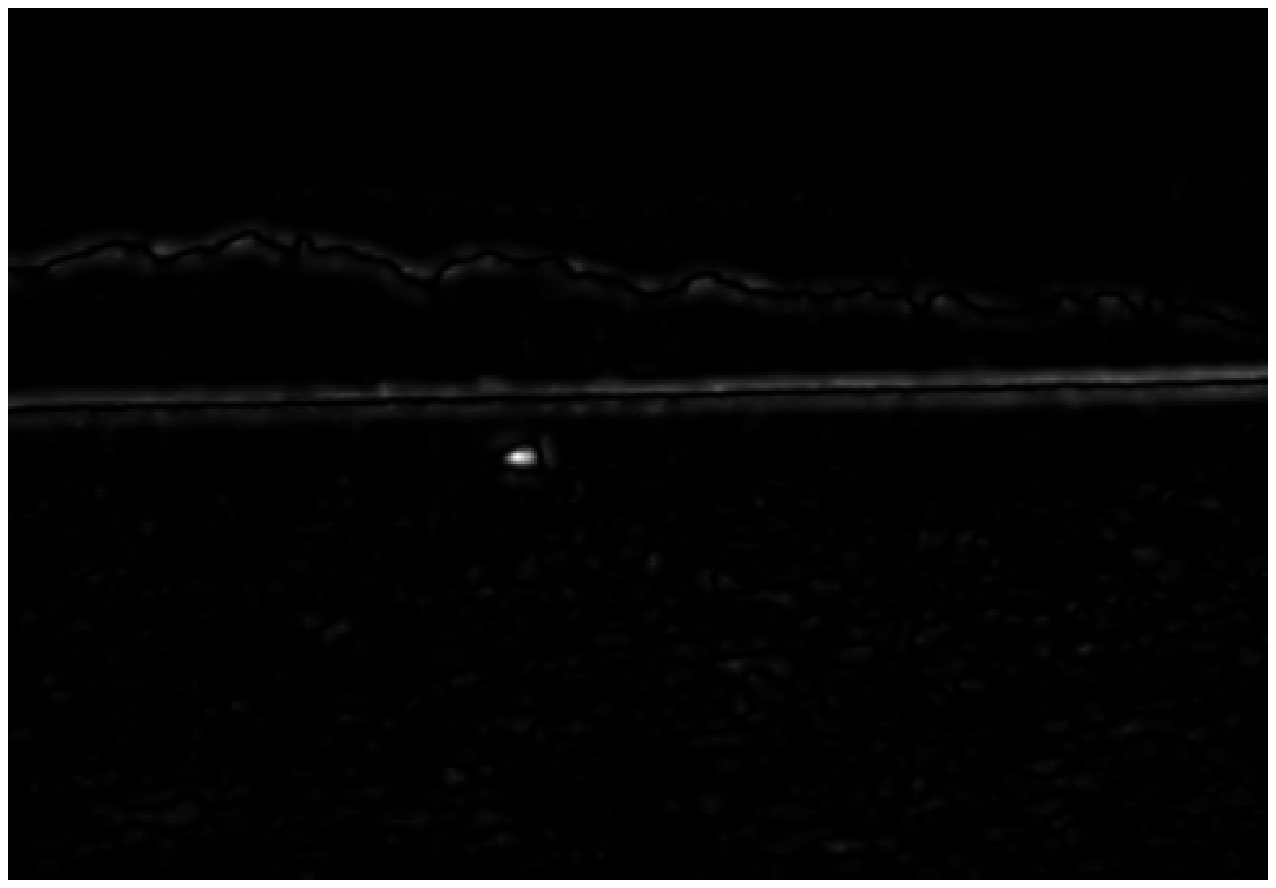}%
		\label{fig:aagd2}}
	~
	\subfloat[]{\includegraphics[width=0.93in]{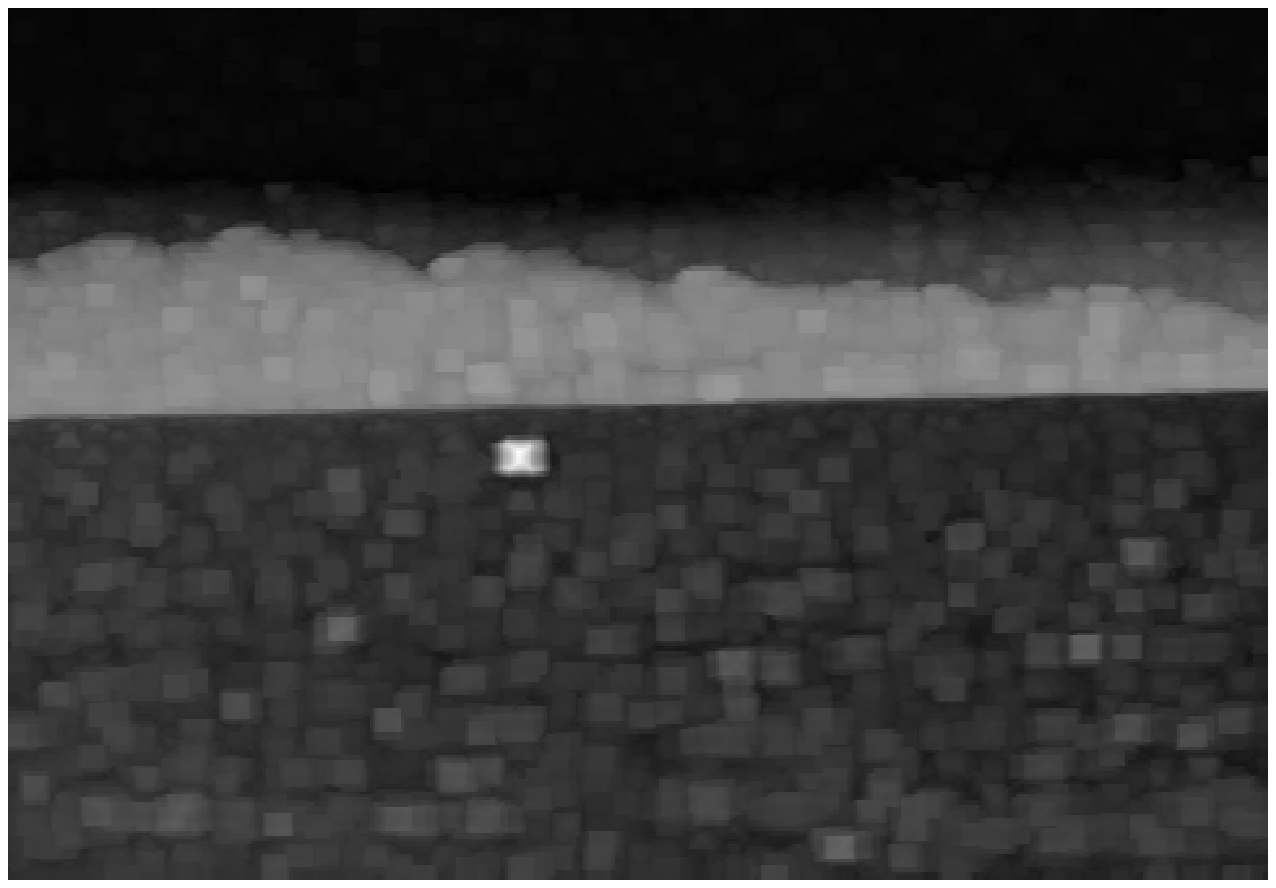}%
		\label{fig:lcm2}}
	~
	\subfloat[]{\includegraphics[width=0.93in]{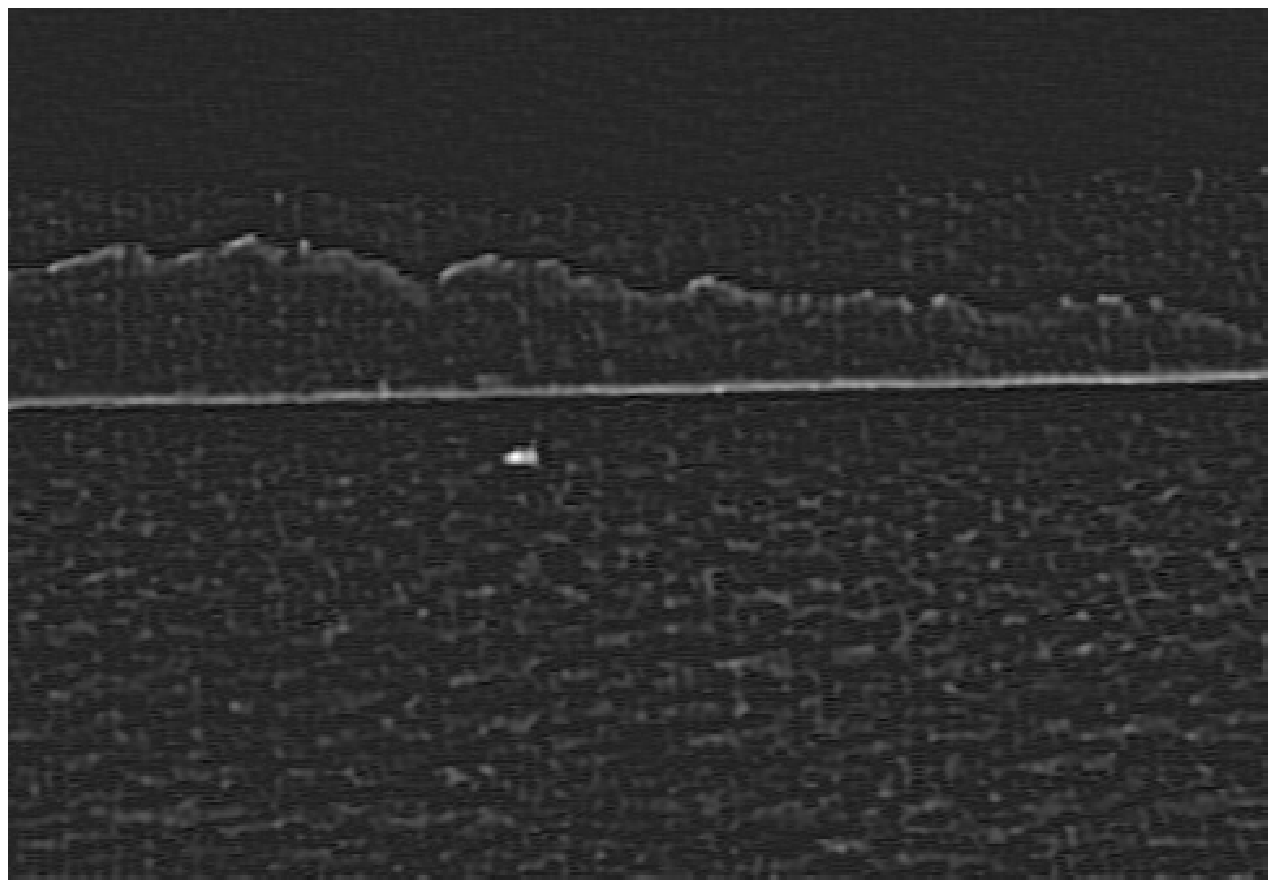}%
		\label{fig:log2}}
	~
	\subfloat[]{\includegraphics[width=0.93in]{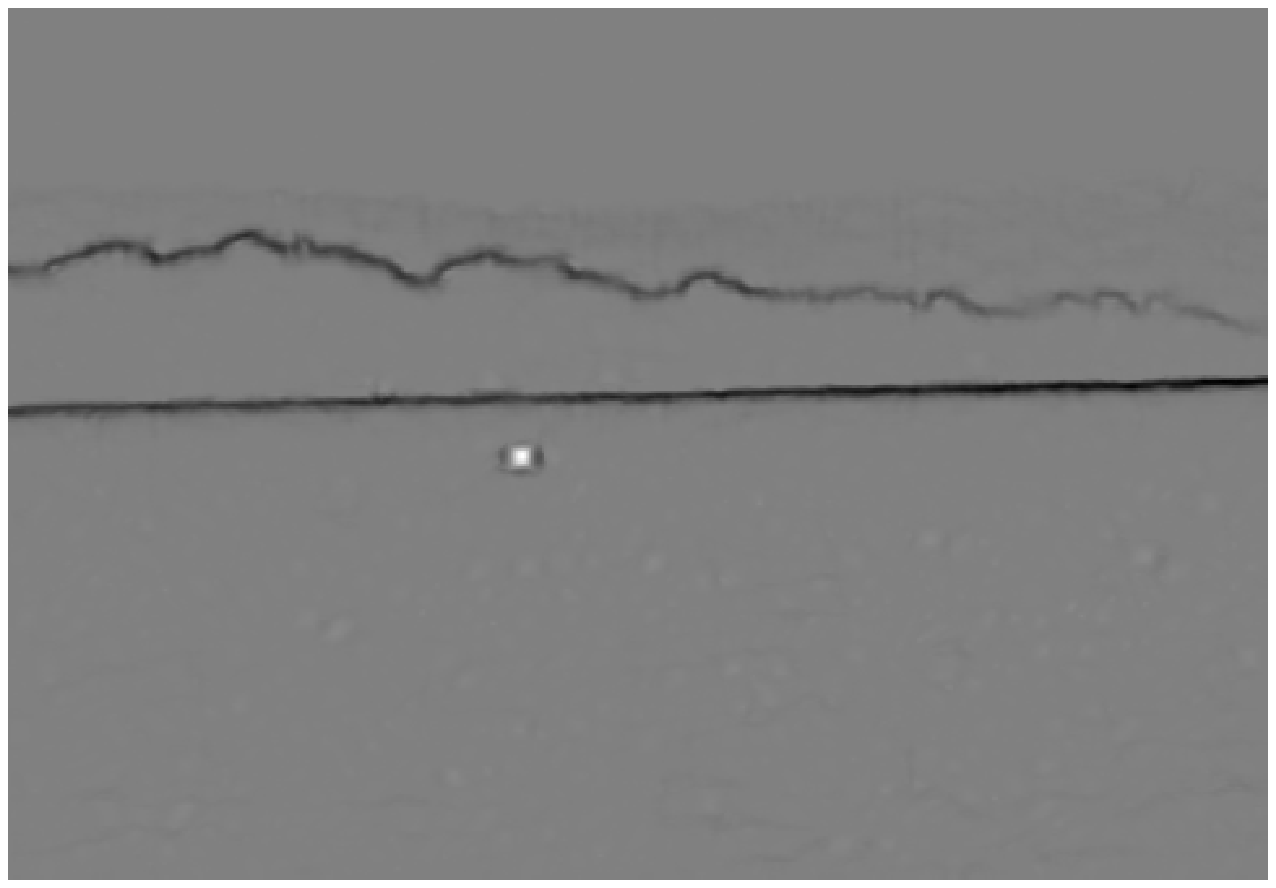}%
		\label{fig:pcm2}}
	~
	\subfloat[]{\includegraphics[width=0.93in]{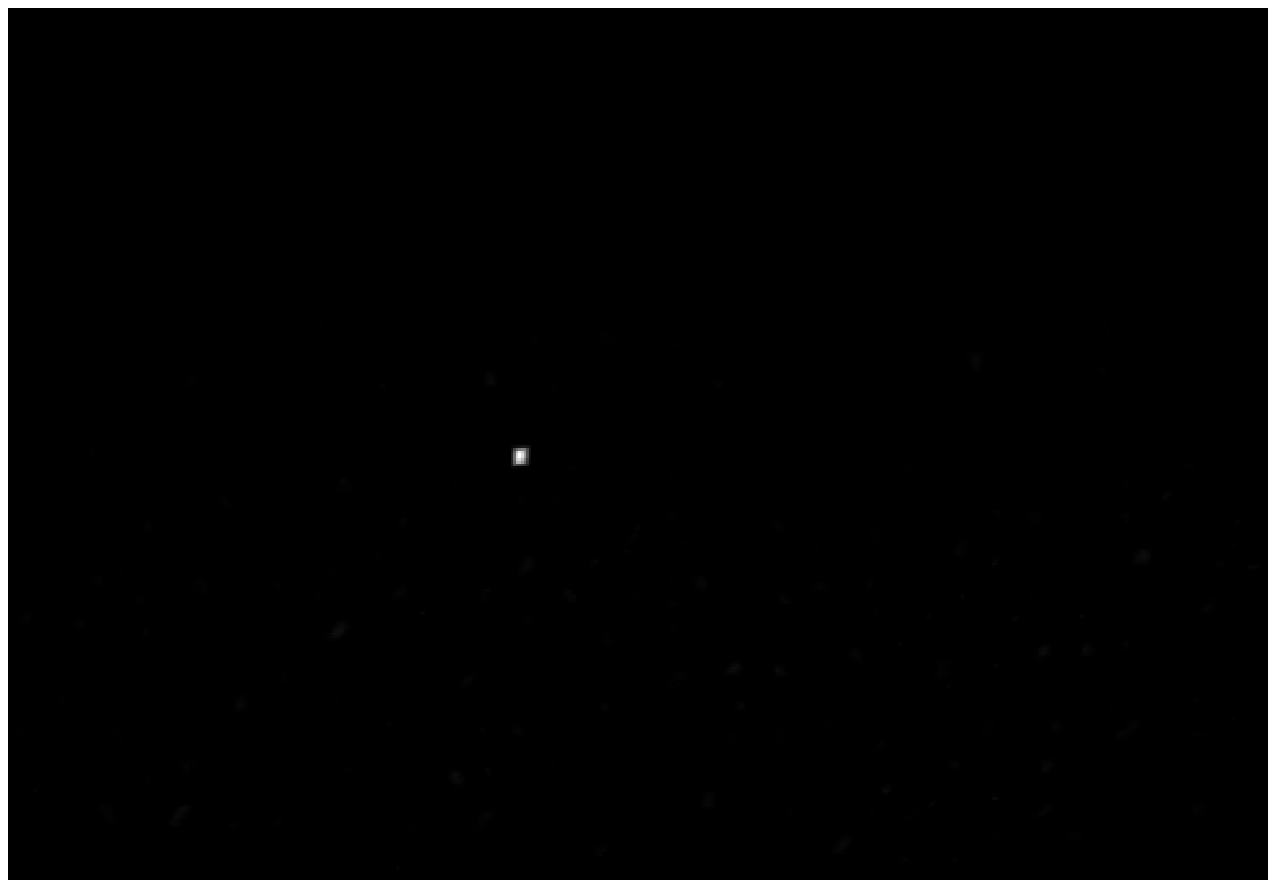}%
		\label{fig:prop2}}
		
			\subfloat[]{\includegraphics[width=0.93in]{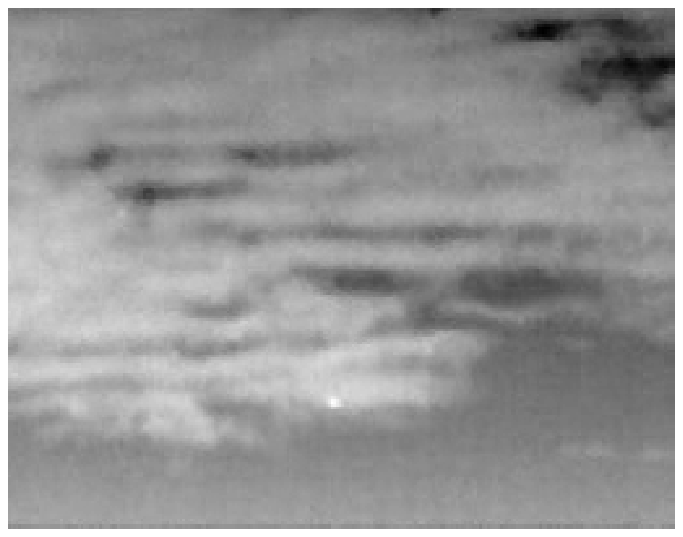}%
		\label{fig:original3}}
		~
		\subfloat[]{\includegraphics[width=0.93in]{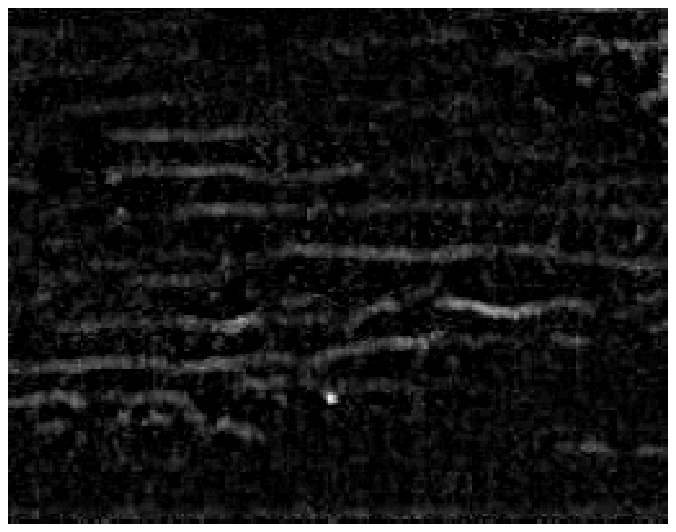}%
		\label{fig:top3}}
	~
	\subfloat[]{\includegraphics[width=0.93in]{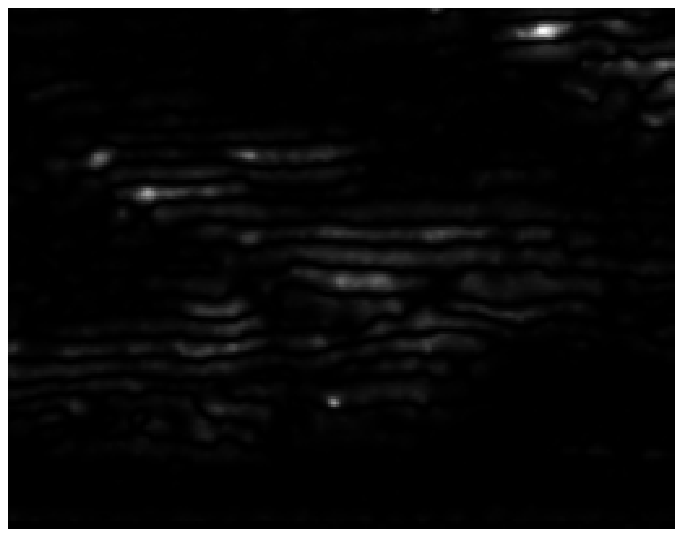}%
		\label{fig:aagd3}}
	~
	\subfloat[]{\includegraphics[width=0.93in]{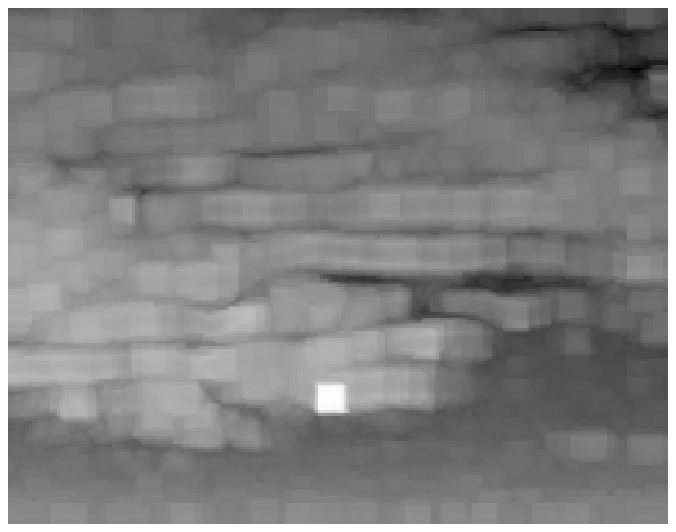}%
		\label{fig:lcm3}}
	~
	\subfloat[]{\includegraphics[width=0.93in]{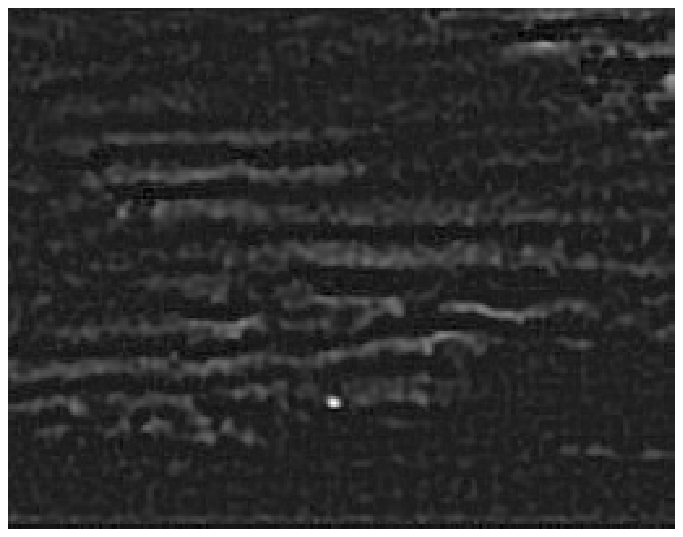}%
		\label{fig:log3}}
	~
	\subfloat[]{\includegraphics[width=0.93in]{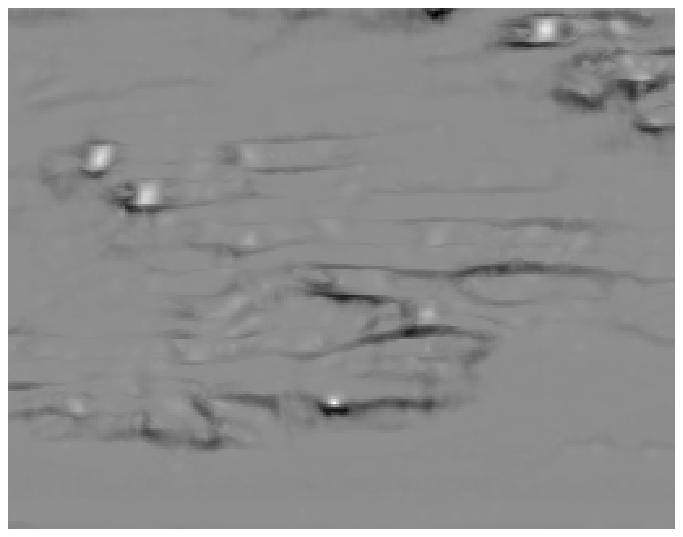}%
		\label{fig:pcm3}}
	~
	\subfloat[]{\includegraphics[width=0.93in]{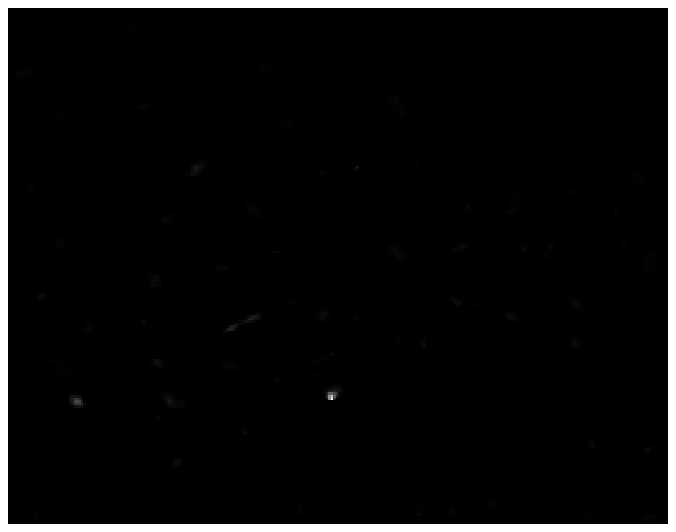}%
		\label{fig:prop3}}
		
			\subfloat[]{\includegraphics[width=0.93in]{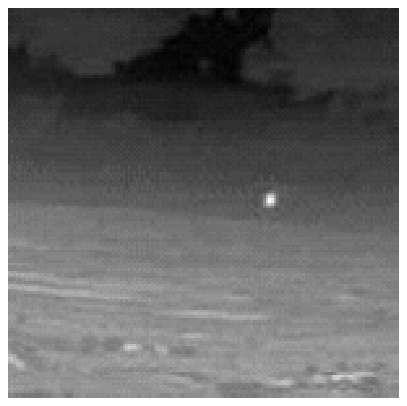}%
		\label{fig:original5}}
		~
		\subfloat[]{\includegraphics[width=0.93in]{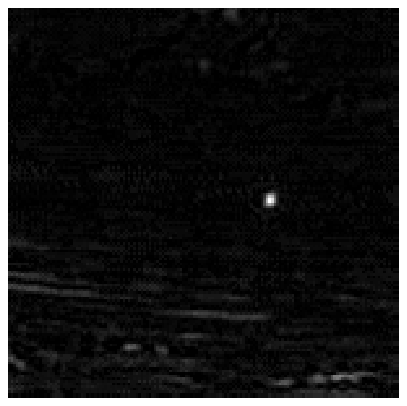}%
		\label{fig:top5}}
	~
	\subfloat[]{\includegraphics[width=0.93in]{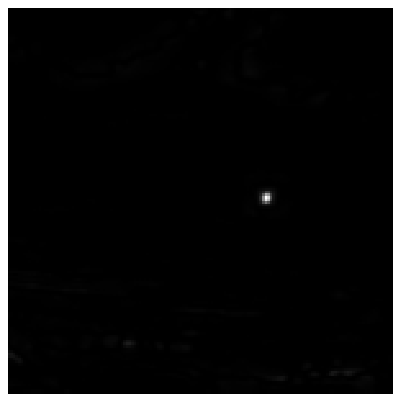}%
		\label{fig:aagd5}}
	~
	\subfloat[]{\includegraphics[width=0.93in]{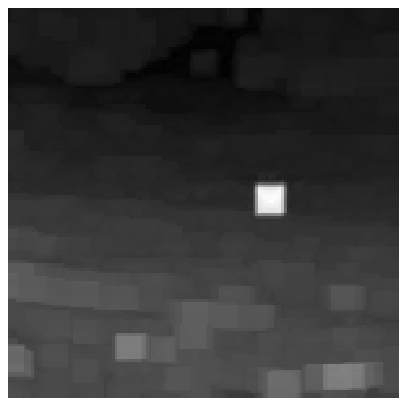}%
		\label{fig:lcm5}}
	~
	\subfloat[]{\includegraphics[width=0.93in]{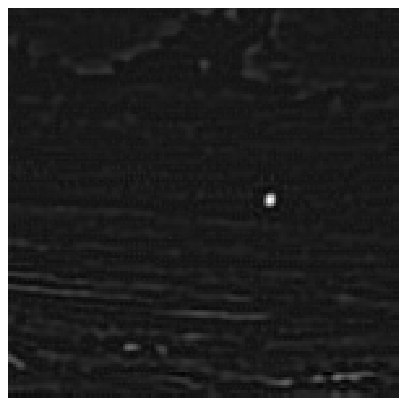}%
		\label{fig:log5}}
	~
	\subfloat[]{\includegraphics[width=0.93in]{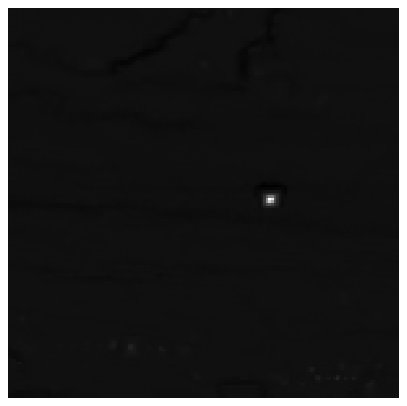}%
		\label{fig:pcm5}}
	~
	\subfloat[]{\includegraphics[width=0.93in]{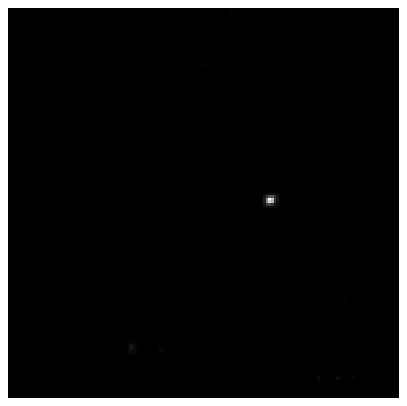}%
		\label{fig:prop5}}
		
			\subfloat[]{\includegraphics[width=0.93in]{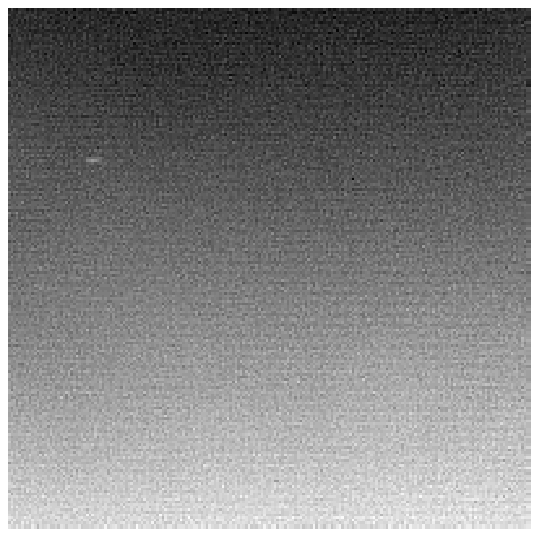}%
		\label{fig:original6}}
		~
		\subfloat[]{\includegraphics[width=0.93in]{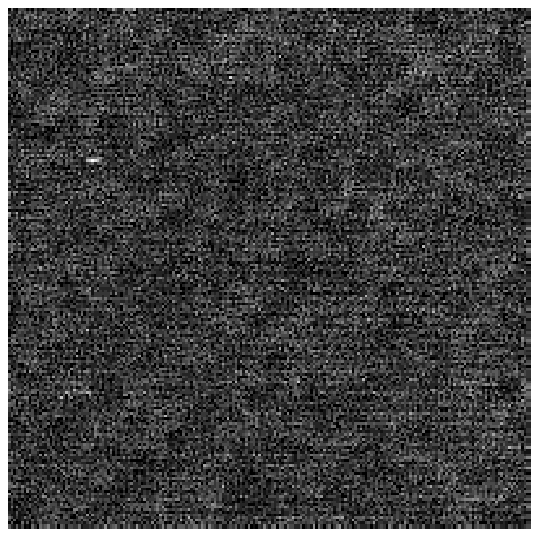}%
		\label{fig:top6}}
	~
	\subfloat[]{\includegraphics[width=0.93in]{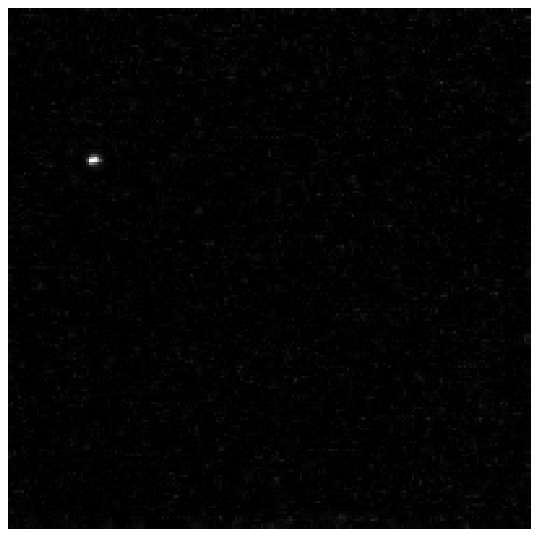}%
		\label{fig:aagd6}}
	~
	\subfloat[]{\includegraphics[width=0.93in]{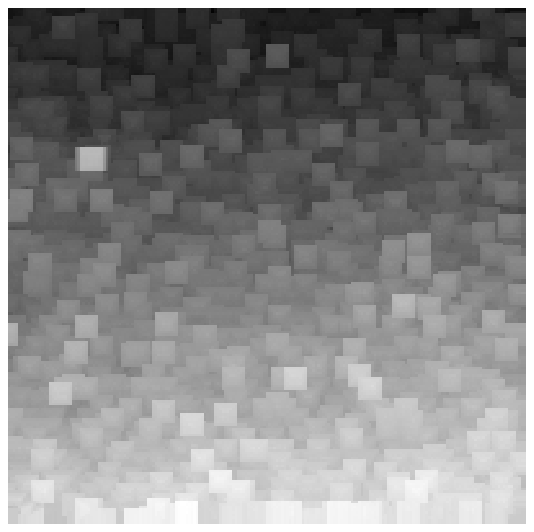}%
		\label{fig:lcm6}}
	~
	\subfloat[]{\includegraphics[width=0.93in]{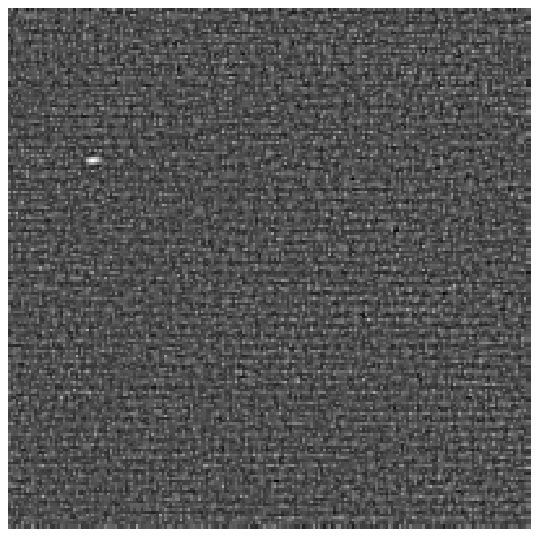}%
		\label{fig:log6}}
	~
	\subfloat[]{\includegraphics[width=0.93in]{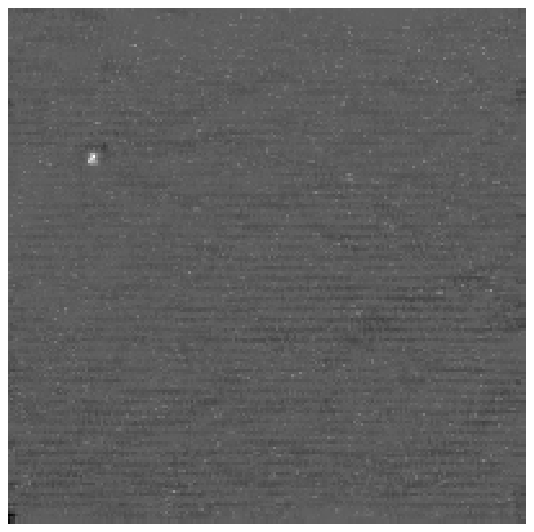}%
		\label{fig:pcm6}}
	~
	\subfloat[]{\includegraphics[width=0.93in]{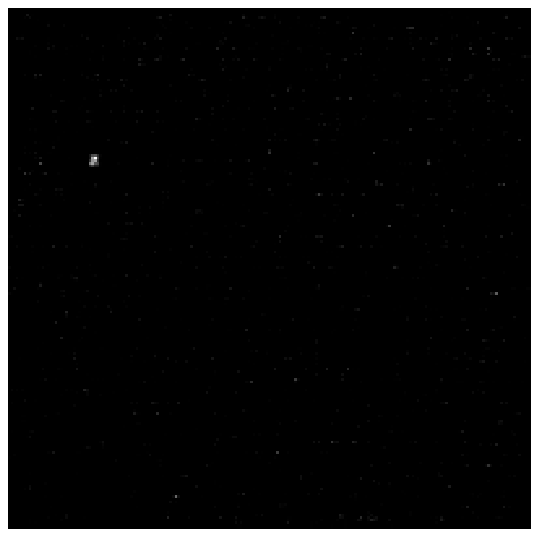}%
		\label{fig:prop6}}
		
					\subfloat[]{\includegraphics[width=0.93in]{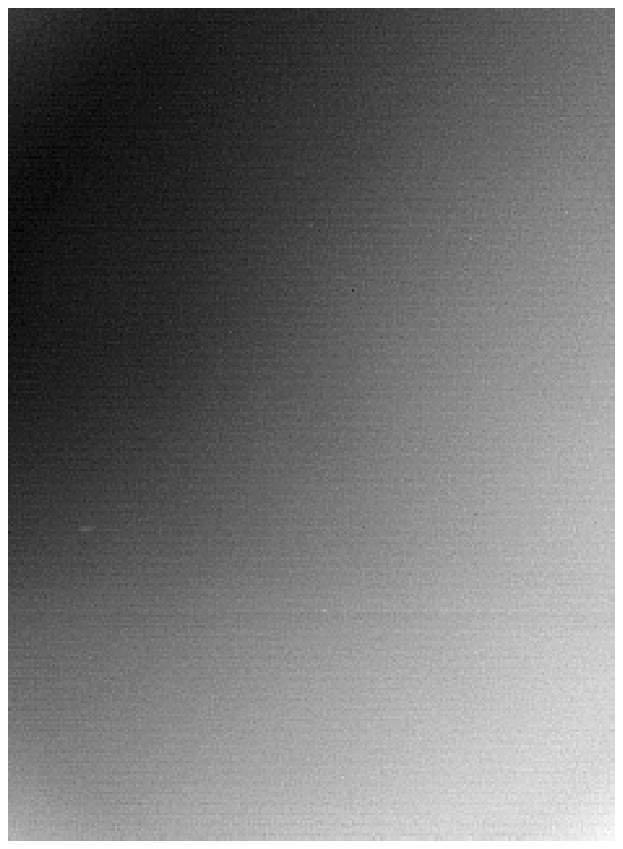}%
		\label{fig:original11}}
		~
		\subfloat[]{\includegraphics[width=0.93in]{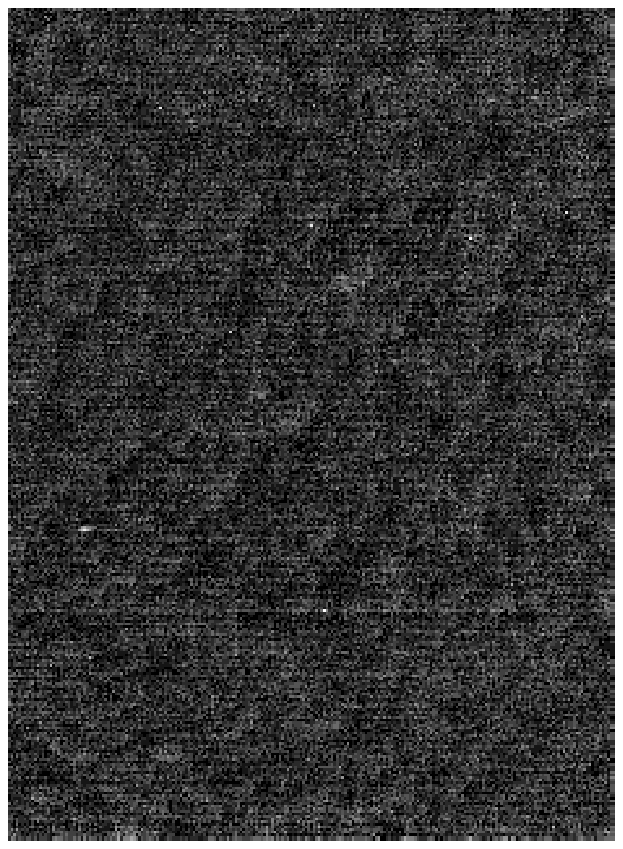}%
		\label{fig:top11}}
	~
	\subfloat[]{\includegraphics[width=0.93in]{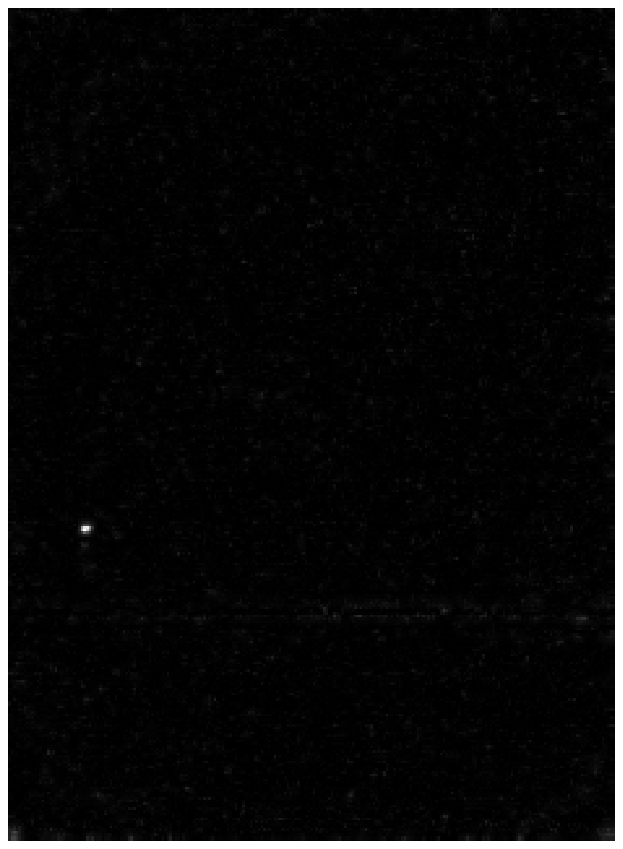}%
		\label{fig:aagd11}}
	~
	\subfloat[]{\includegraphics[width=0.93in]{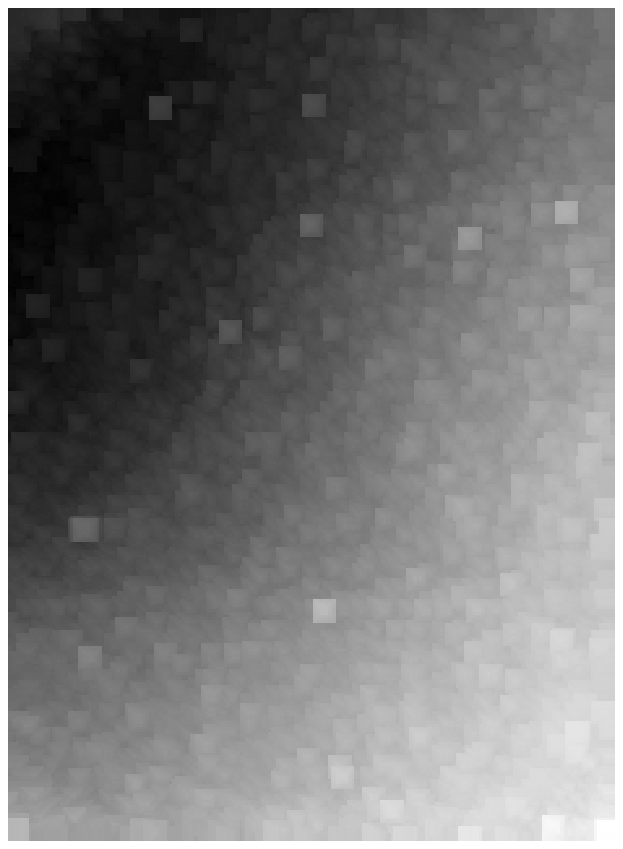}%
		\label{fig:lcm11}}
	~
	\subfloat[]{\includegraphics[width=0.93in]{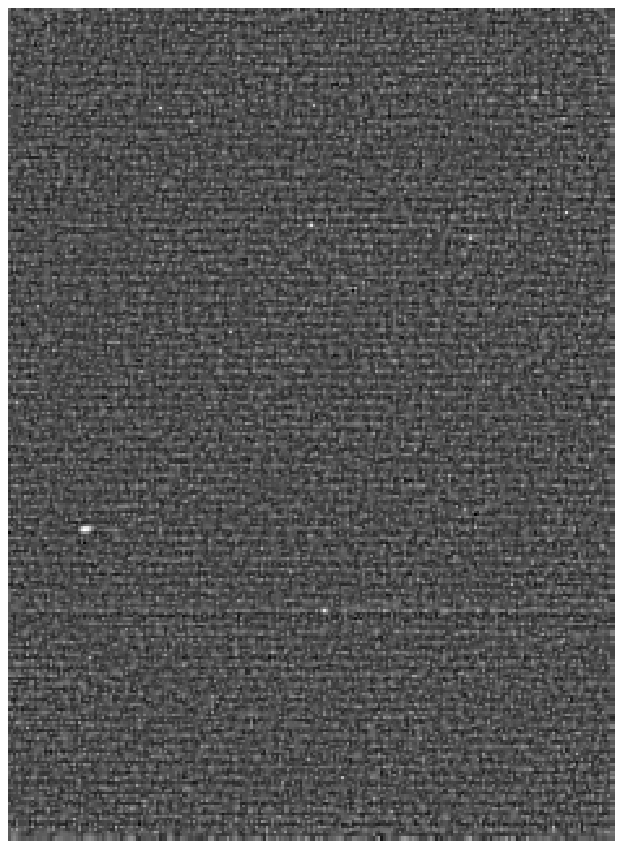}%
		\label{fig:log11}}
	~
	\subfloat[]{\includegraphics[width=0.93in]{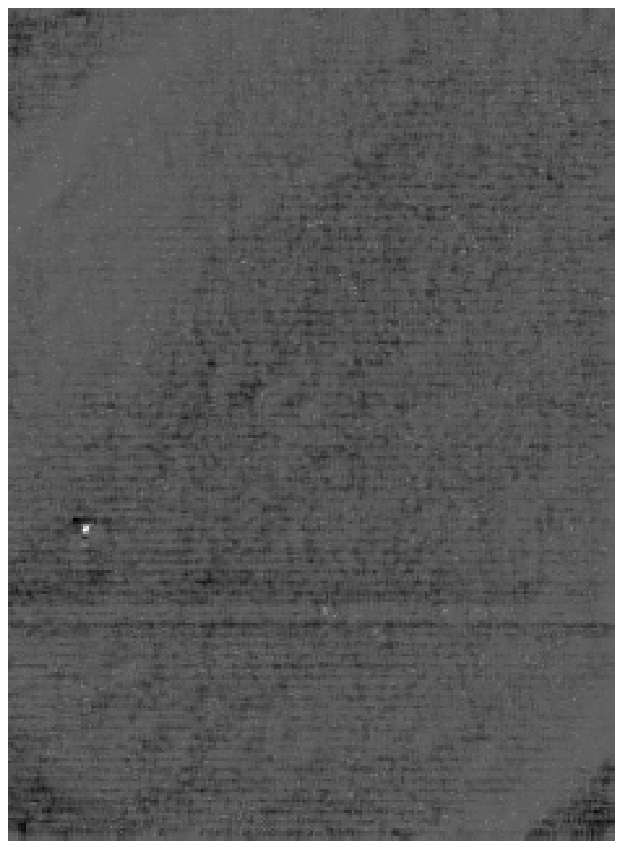}%
		\label{fig:pcm11}}
	~
	\subfloat[]{\includegraphics[width=0.93in]{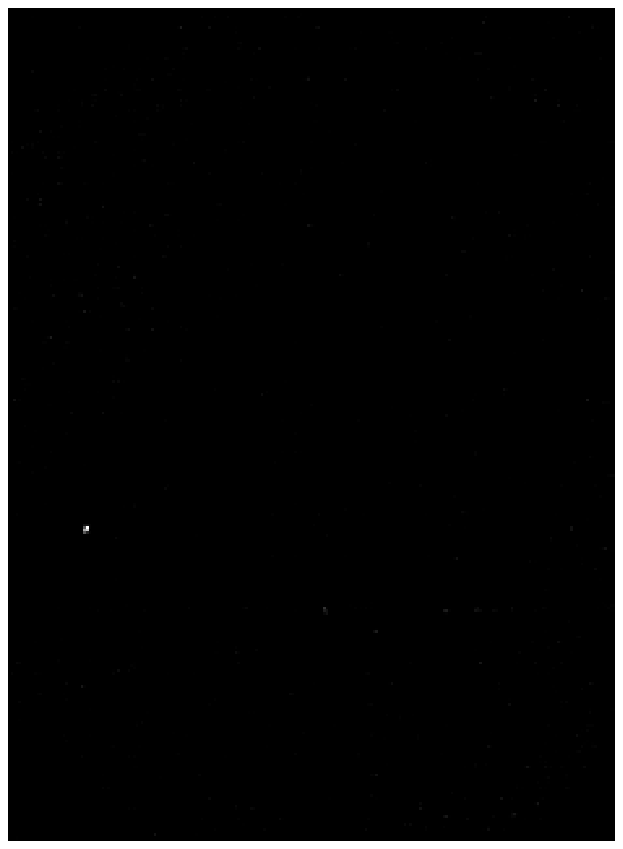}%
		\label{fig:prop11}}

					\subfloat[]{\includegraphics[width=0.93in]{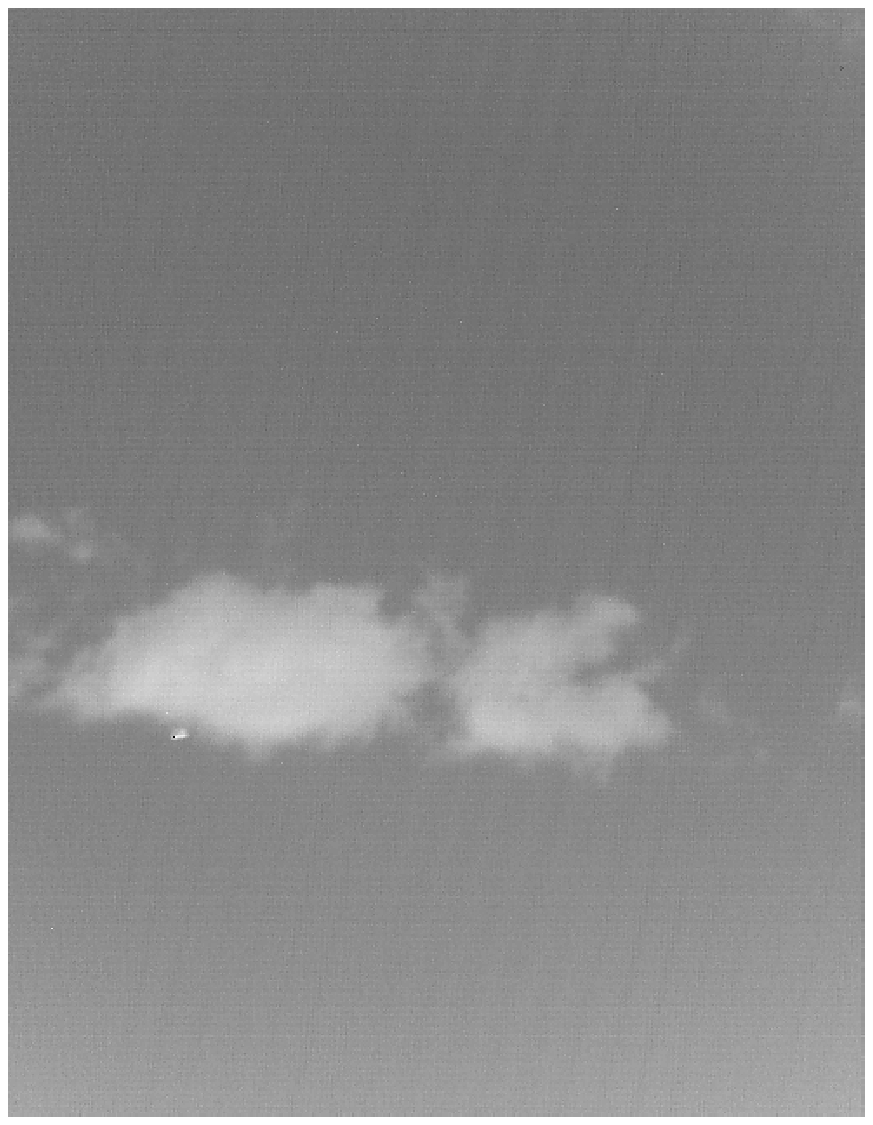}%
		\label{fig:original15}}
		~
		\subfloat[]{\includegraphics[width=0.93in]{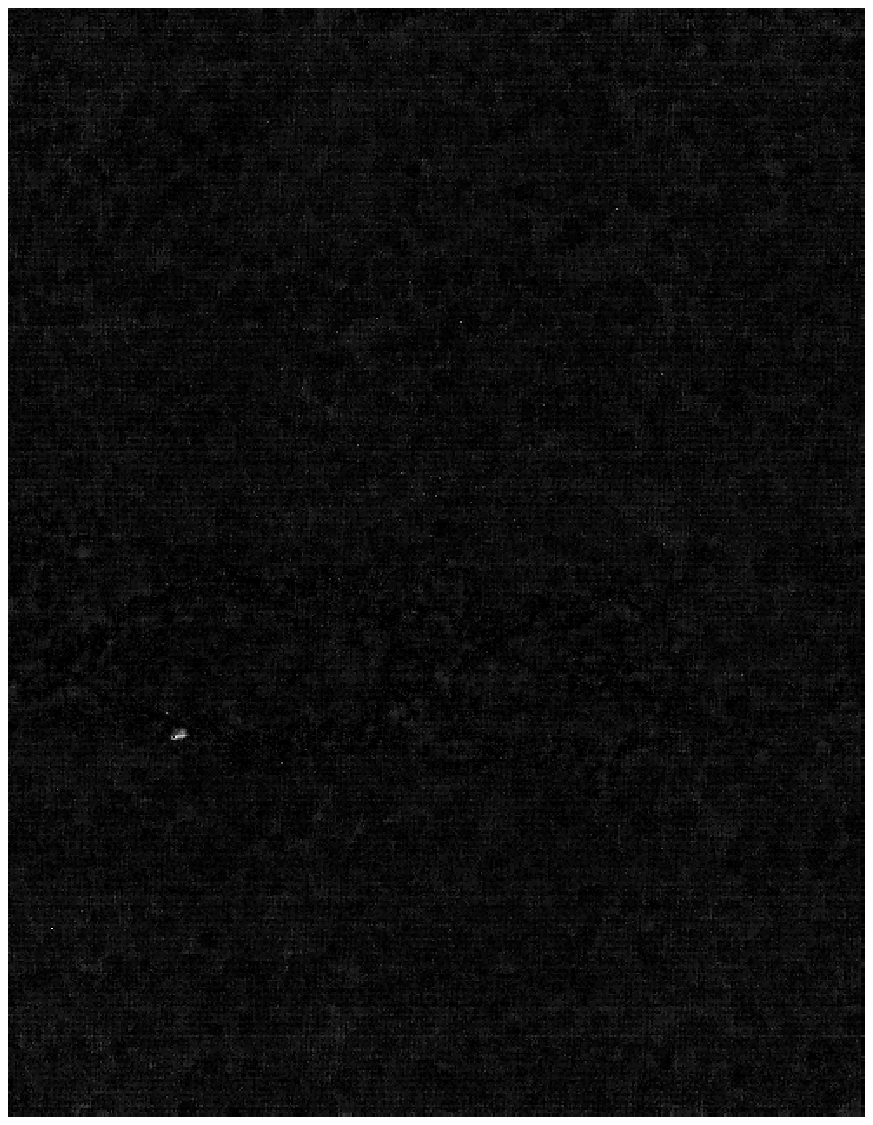}%
		\label{fig:top15}}
	~
	\subfloat[]{\includegraphics[width=0.93in]{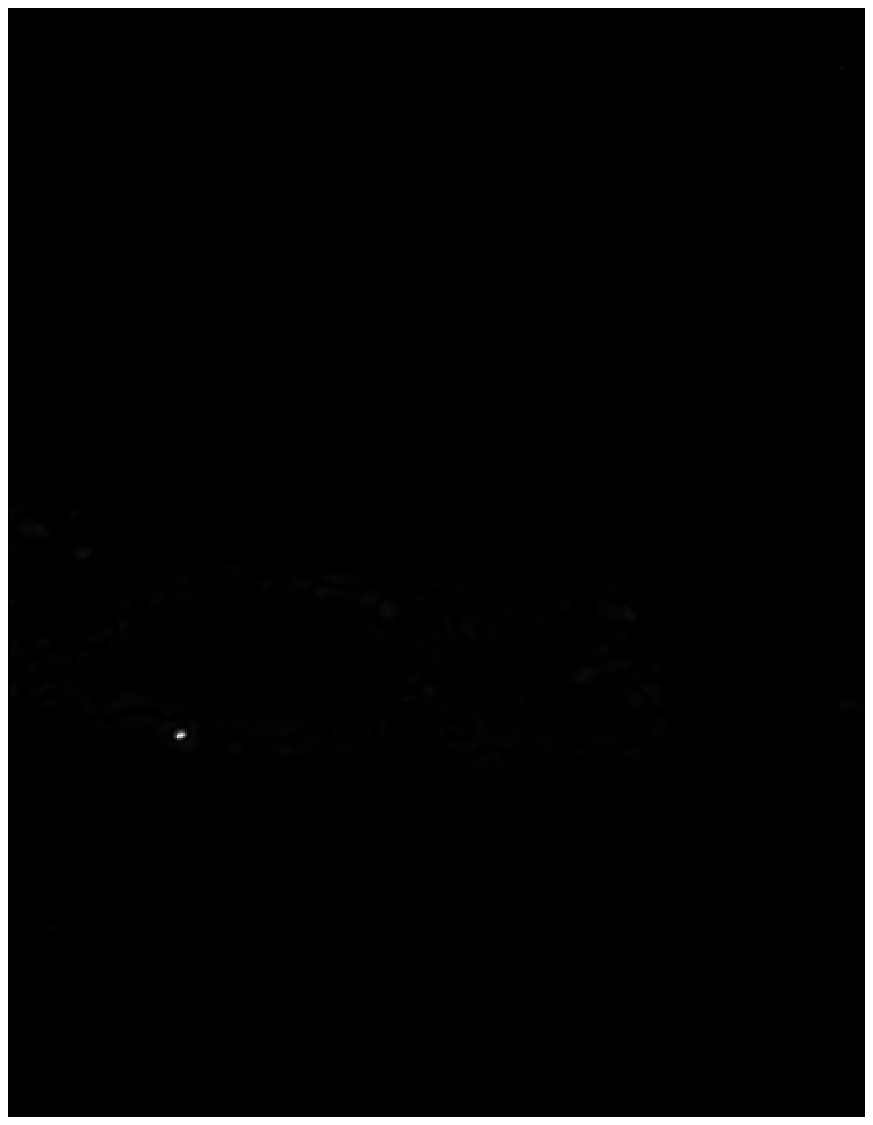}%
		\label{fig:aagd15}}
	~
	\subfloat[]{\includegraphics[width=0.93in]{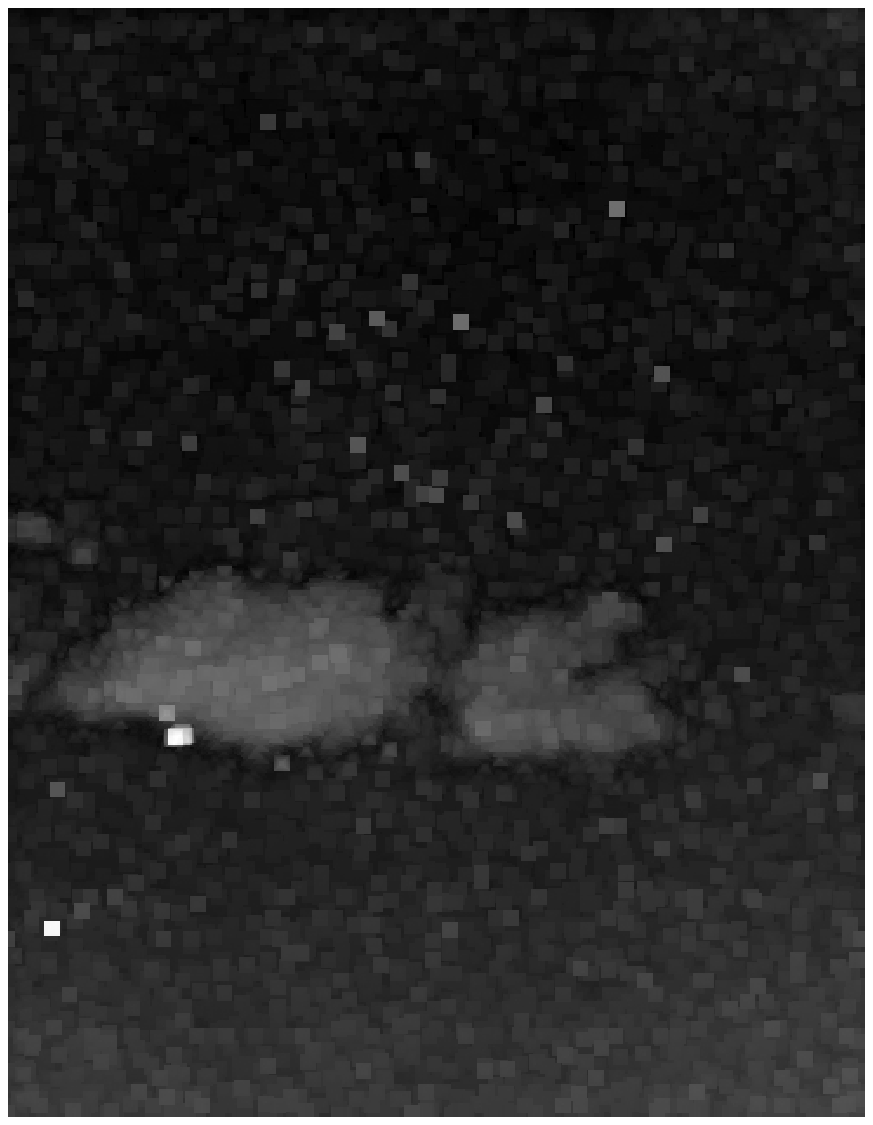}%
		\label{fig:lcm15}}
	~
	\subfloat[]{\includegraphics[width=0.93in]{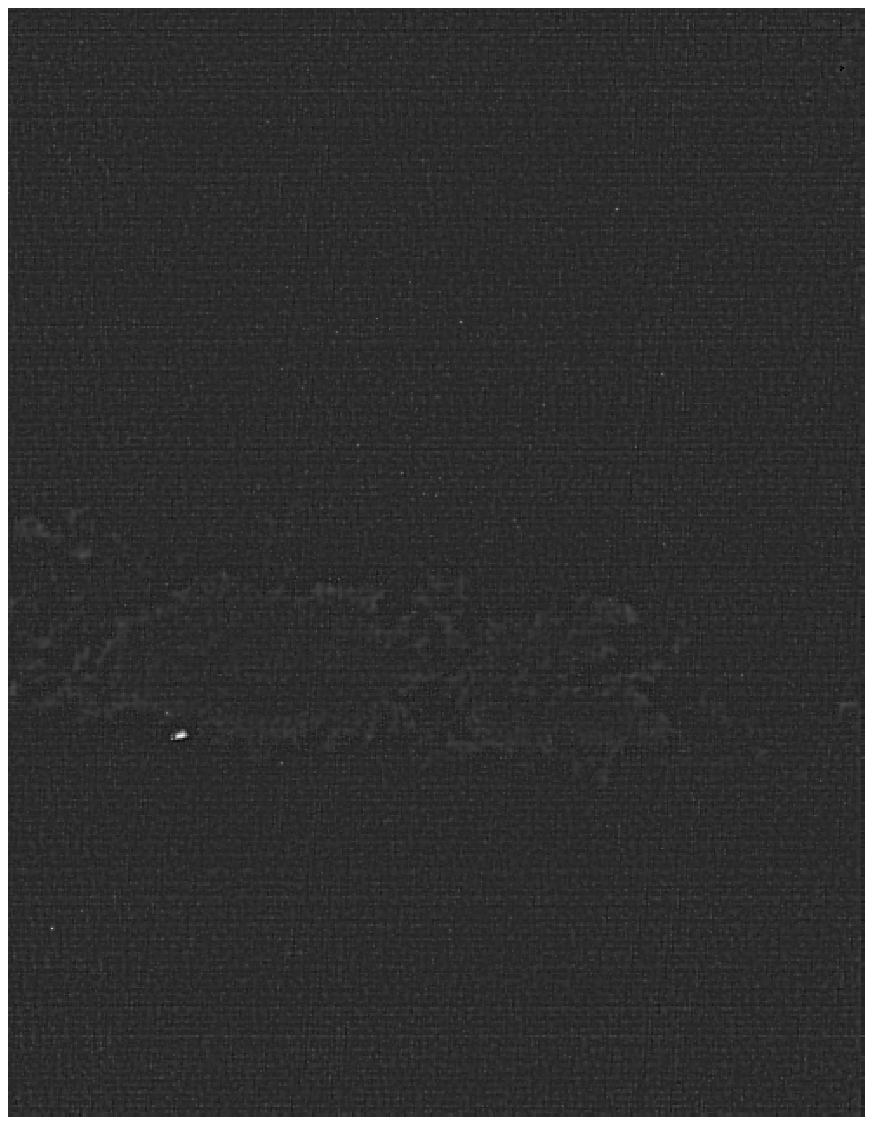}%
		\label{fig:log15}}
	~
	\subfloat[]{\includegraphics[width=0.93in]{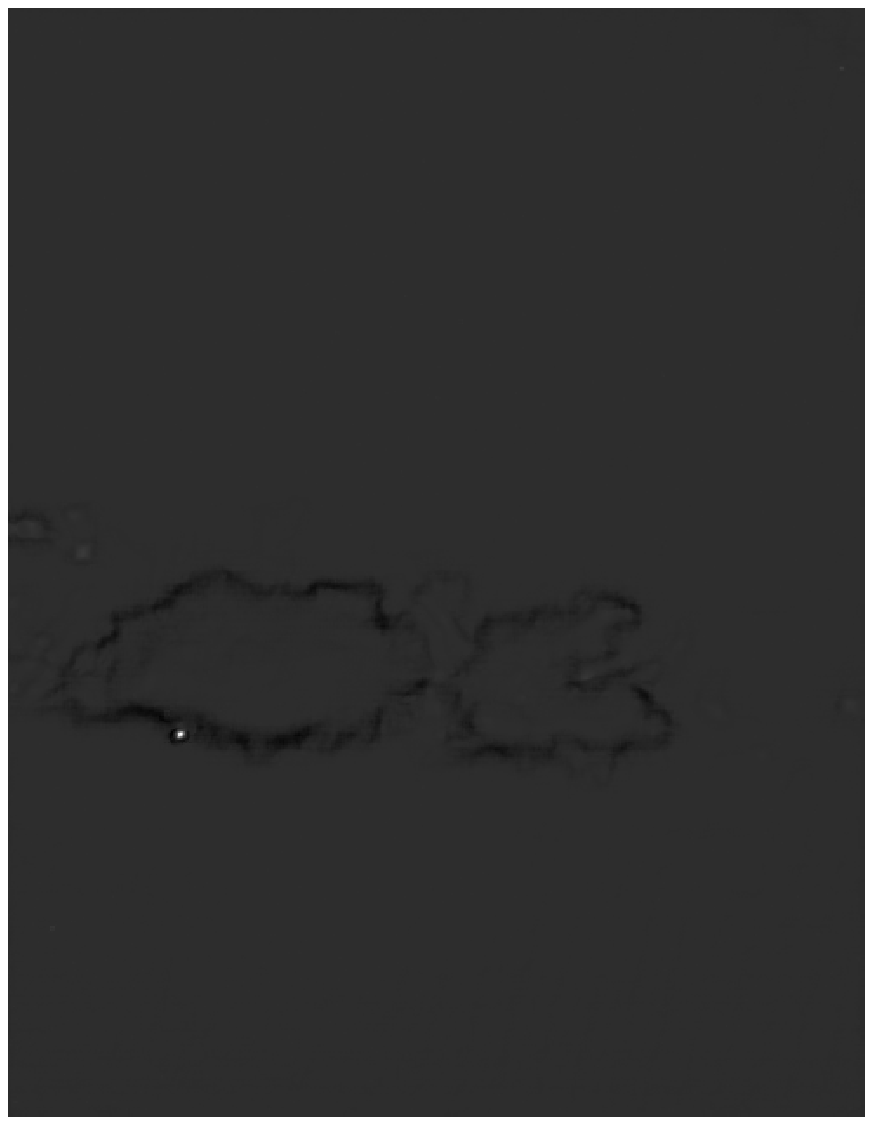}%
		\label{fig:pcm15}}
	~
	\subfloat[]{\includegraphics[width=0.93in]{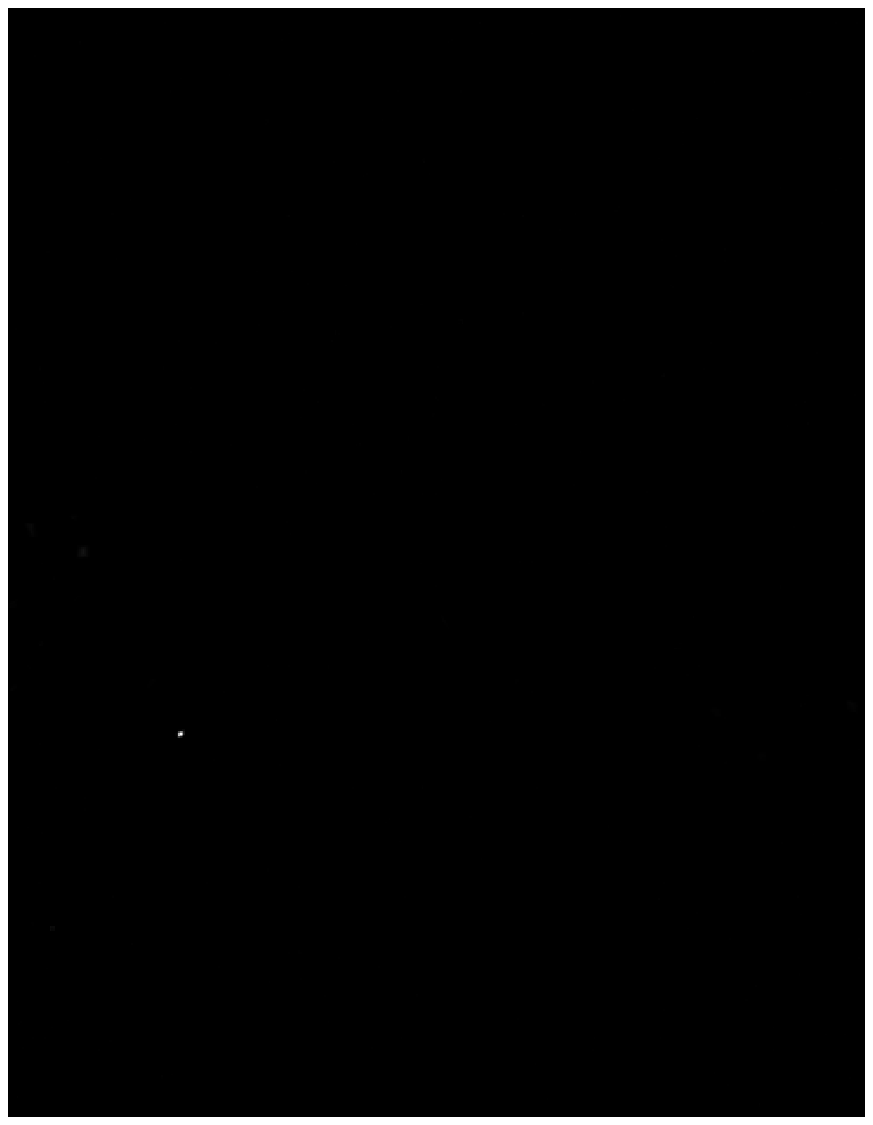}%
		\label{fig:prop15}}
		\caption{pre-thresholding filtering results on the infrared images. {\bf a, h, o, v, ac, aj)} original images , {\bf b, i, p, w, ad, ak)} filtering results of Top-Hat,  {\bf c, j, q, x, ae, al)} filtering results of AAGD, {\bf d, k, r, y, af, am)} filtering results of LCM, {\bf e, l, s, z, ag, an)} filtering results of LoG, {\bf f, m, t, aa, ah , ao)} filtering results of PCM, {\bf g, n, u, ab, ai , ap)} filtering results of \textbf{the proposed algorithm (ADMD)}.}
	\label{fig:common_test}
\end{figure*}
\begin{figure*}[!t]
	\centering
	\subfloat[]{\includegraphics[width=2.2in]{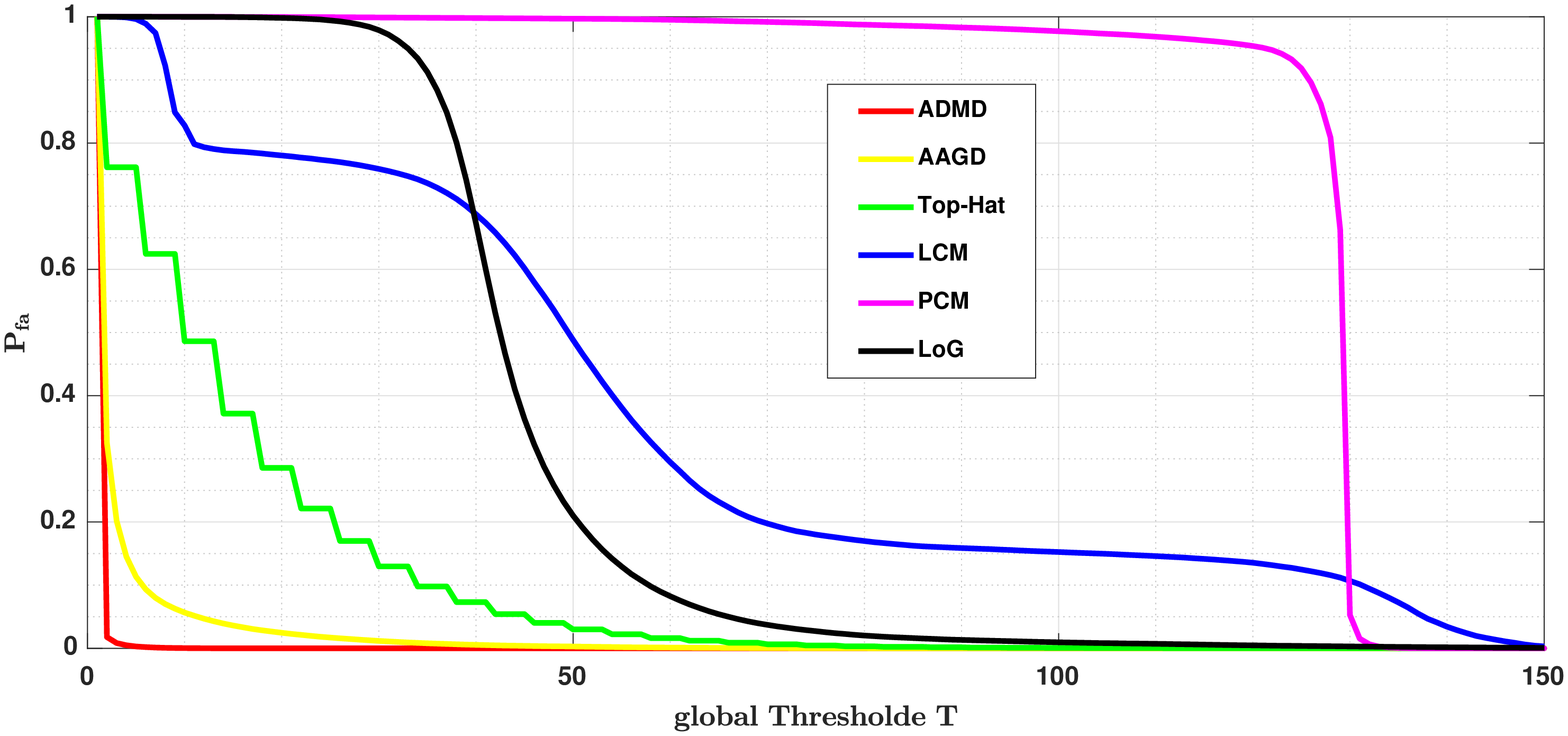}%
		\label{fig:roc1}}
	~
	\subfloat[]{\includegraphics[width=2.2in]{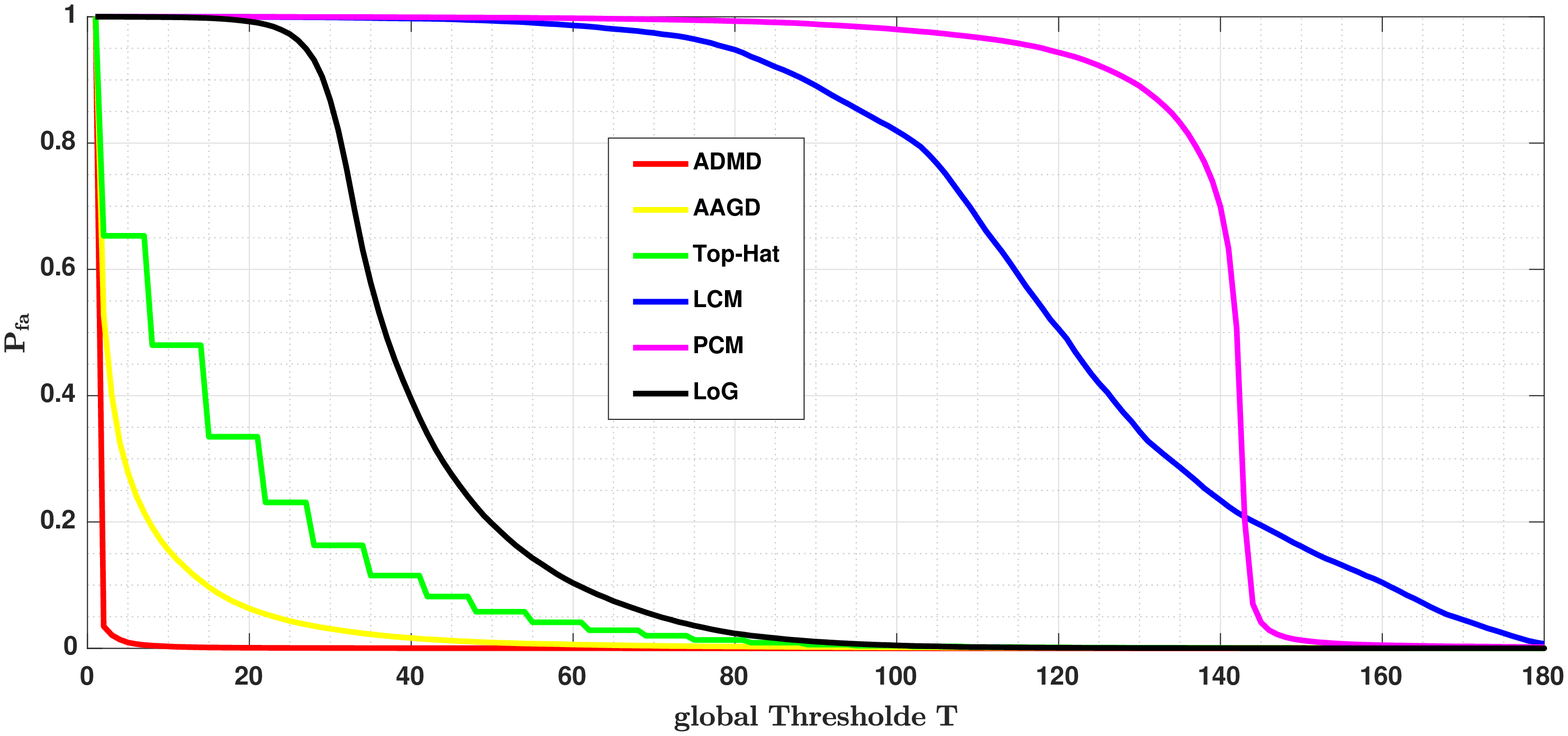}%
		\label{fig:roc2}}
	~
	\subfloat[]{\includegraphics[width=2.2in]{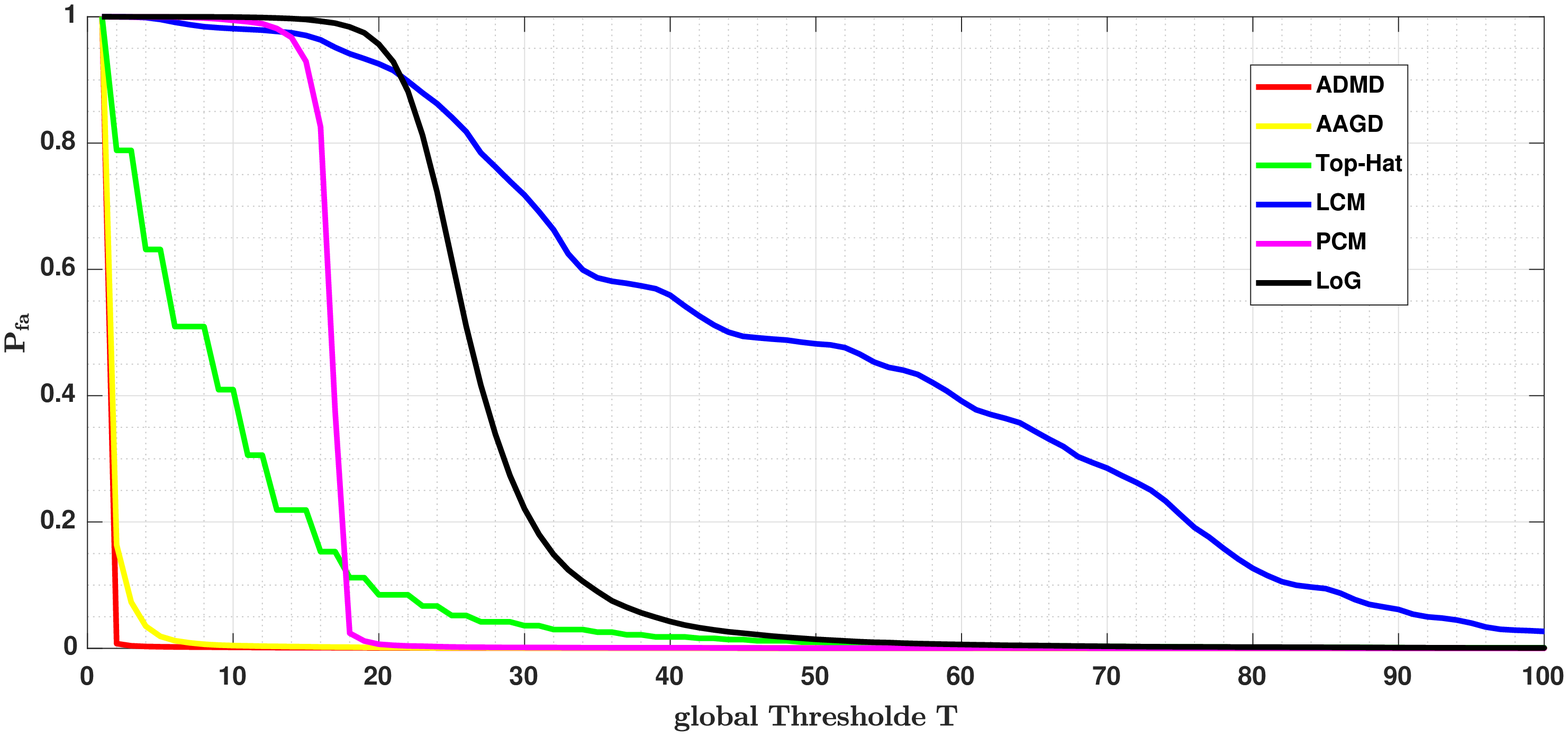}%
		\label{fig:roc3}}
	
	\subfloat[]{\includegraphics[width=2.2in]{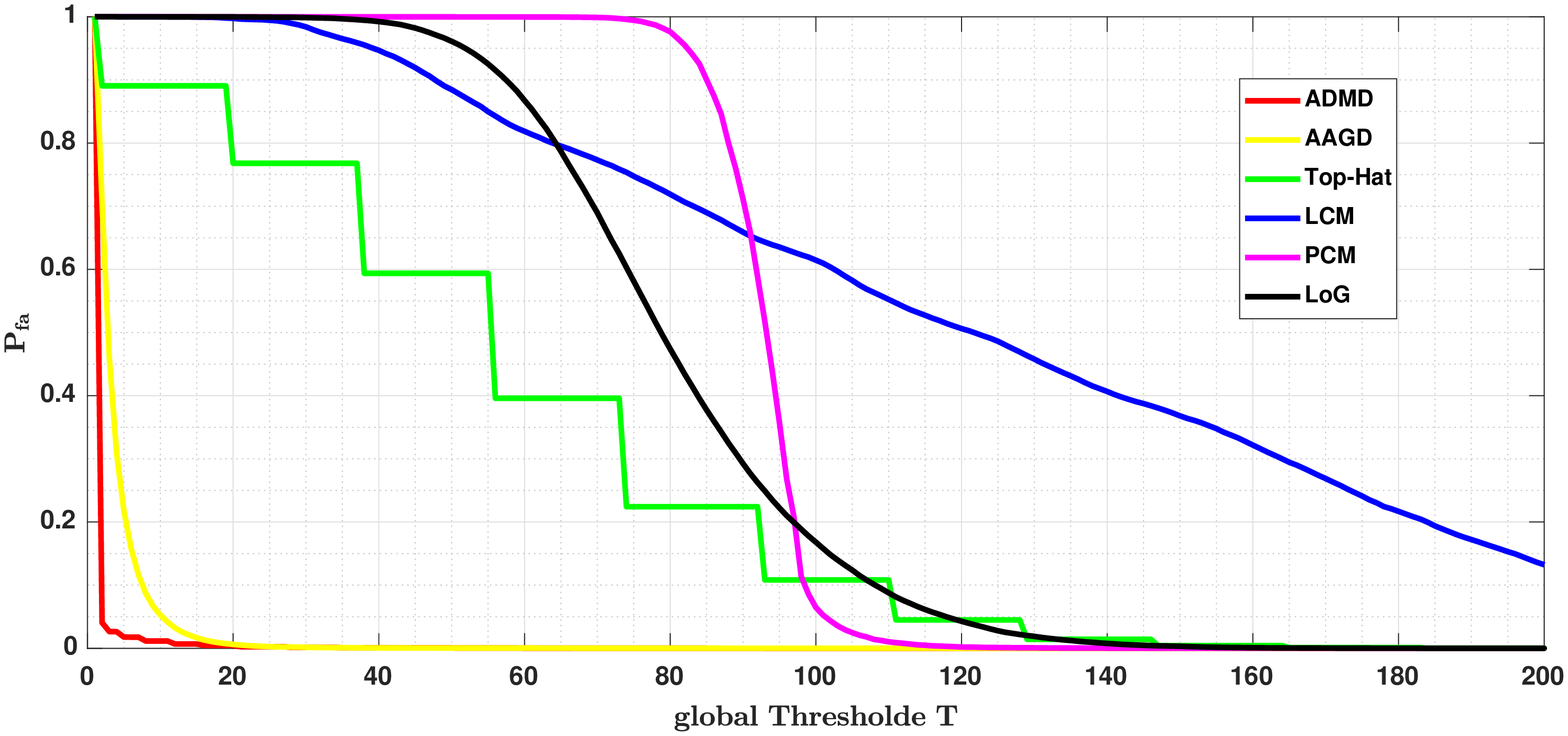}%
	\label{fig:roc4}}
	~
	\subfloat[]{\includegraphics[width=2.2in]{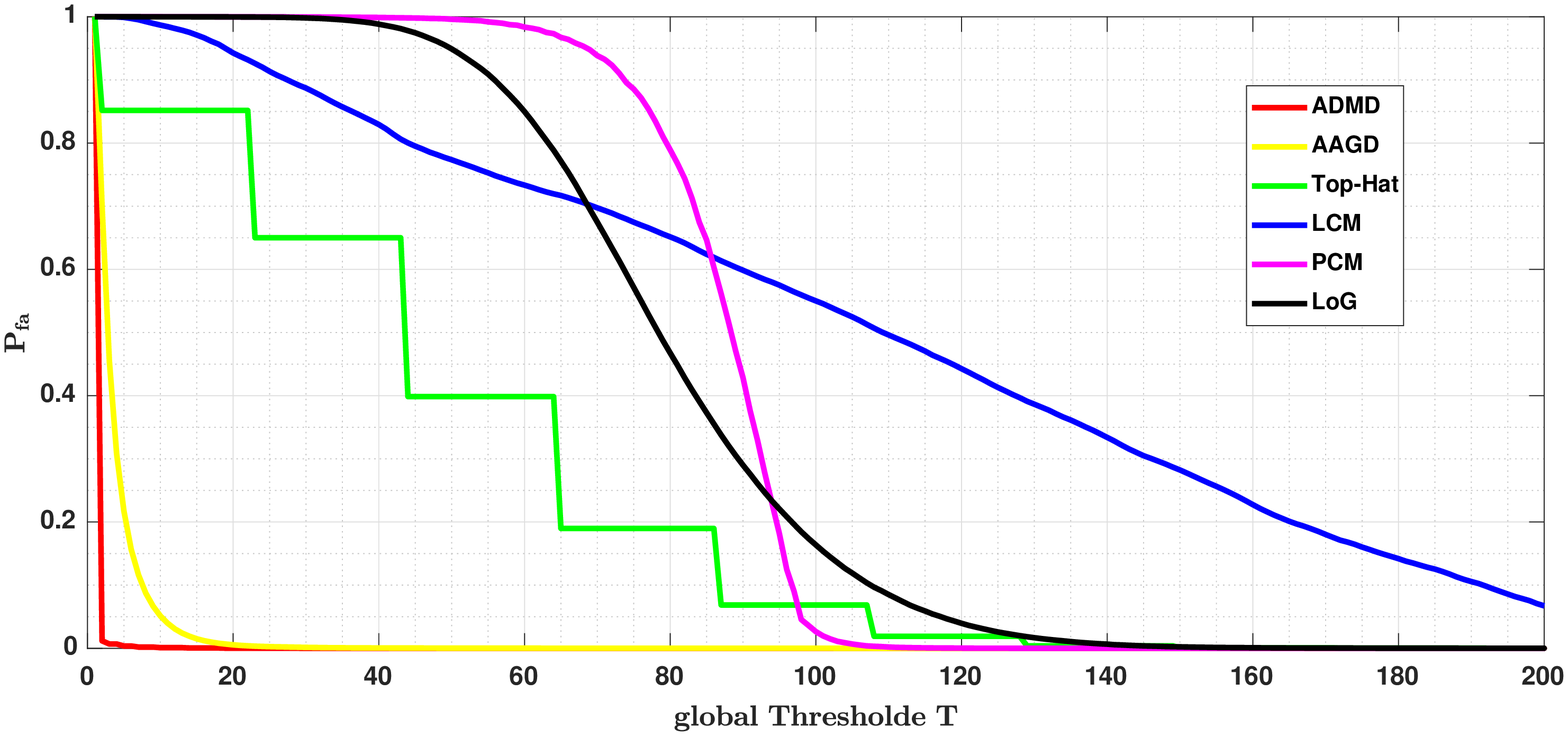}%
		\label{fig:roc5}}
	~
	\subfloat[]{\includegraphics[width=2.2in]{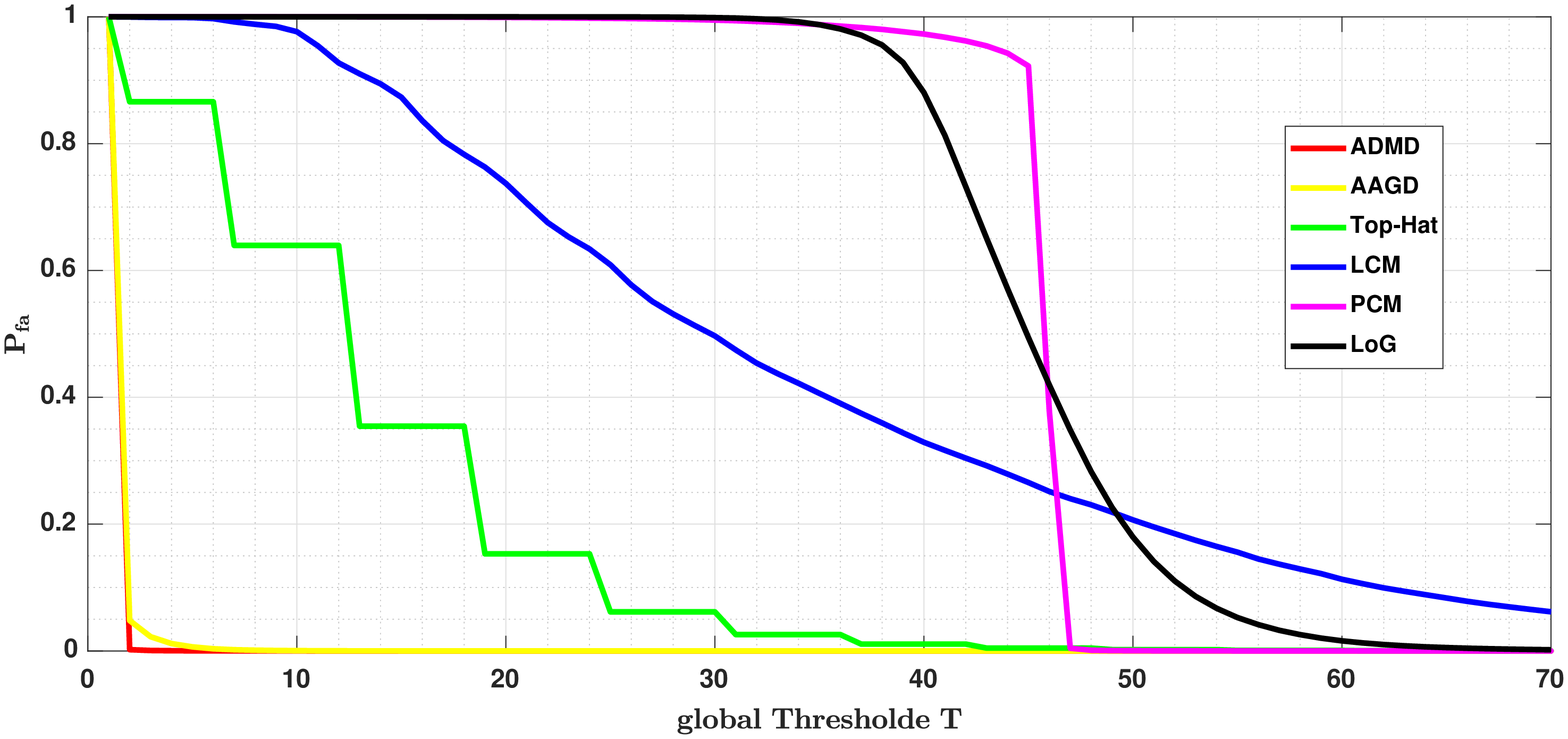}%
		\label{fig:roc6}}
	
	\caption{false alarm rate versus threshold level curves for the different test images including: {\bf a)} \autoref{fig:original2}, {\bf b)} \autoref{fig:original3},  {\bf c)} \autoref{fig:original5}, {\bf d)} \autoref{fig:original6}, {\bf e)} \autoref{fig:original11}  and {\bf f)} \autoref{fig:original15}.}
	\label{fig:roc_test}
\end{figure*}
\setlength\tabcolsep{4pt} 
\begin{table}[!b]
	\centering
	\caption{ SCR values for the test targets}
	\begin{tabular}{|c|c|c|c|c|c|c|}
		\hline
		 &AAGD&TopHat&LCM&PCM&LoG& \textbf {ADMD}\\
		\hline
		\autoref{fig:original2}&4.37&1.94&1.55&3.78&2.43&{\bf 86.23}\\
		\hline
		\autoref{fig:original3}&9.07&7.17&1.78&3.01&7.27&{\bf 22.85}\\
		\hline
		\autoref{fig:original5}&17.31&10.62&2.32&17.14&12.56&{\bf 22.65}\\
		\hline
		\autoref{fig:original6}&11.59&3.14&2.29&8.09&5.25&{\bf 11.96}\\
		\hline
		\autoref{fig:original11}&10.87&2.46&2.25&7.34& 4.59&{\bf 197.28}\\
		\hline
		\autoref{fig:original15}&13.69&5.22&1.66&6.23&6.95&{\bf 549.18}\\
		\hline
		\end{tabular}
	\label{tab:scrgmu}
\end{table}
\setlength\tabcolsep{4pt} 
\begin{table}[!b]
	\centering
	\caption{ BSF values for the test images}
	\begin{tabular}{|c|c|c|c|c|c|c|}
		\hline
		 &AAGD&TopHat&LCM&PCM&LoG&\textbf {ADMD}\\
		\hline
		\autoref{fig:original2} &  8.66    & 3.36   &1.29 &5.41&4.00 & {\bf 114.23} \\
		\hline
		\autoref{fig:original3}  &  1.89  & 1.09  & 0.75&1.77& 1.43& {\bf 14.77} \\
		\hline
		\autoref{fig:original5} &   14.27 &  2.93  &  1.05 &16.06&4.17& {\bf 25.09} \\
		\hline
		\autoref{fig:original6} &  2.36   &  0.26  & 0.16 &1.47& 0.46& {\bf 3.44} \\
		\hline
		\autoref{fig:original11} &   4.88   & 0.55 & 0.29  &1.81& 0.85& {\bf 25.01} \\
		\hline
		\autoref{fig:original15} &   11.59  &  0.89 & 0.38 &3.52& 1.44& {\bf 45.29} \\
		\hline
		\end{tabular}
	\label{tab:bsf}
\end{table}
In order to demonstrate the effectiveness of the proposed small target detection algorithm, simulation results are provided here. To compare detection performances, some basic as well as state of the art algorithms like as Top-Hat \cite{zeng2006design}, Laplacian of Gaussian (LoG) \cite{kim2011robust}, local contrast measure (LCM) \cite{chen2014local}, absolute average gray difference (AAGD) \cite{deng2016infrared}, and patch-based contrast measure (PCM) \cite{wei2016multiscale} are chosen as baseline algorithms. Except for MS-LoG, which uses 12 scales\cite{kim2012scale}, the rest of the algorithms (AAGD, LCM, PCM, and ADMD) utilize four scales. Typically, a small target occupies less than 80 pixels \cite{zhang2019infrared}, therefore the target window in AAGD as well as the cell size in LCM and PCM take {[}$3\times 3$, $5\times 5$, $7\times7$, $9\times 9${]} values to cover different size of the small target.
 The test images include naval and aerial infrared targets embedded in complicated backgrounds (\autoref{fig:common_test}). The filtering results of the baseline algorithms  as well as the proposed one are illustrated in \autoref{fig:common_test}. As shown in the figure, the proposed algorithm (ADMD) effectively suppresses background clutter and enhances the small infrared target while the other methods show poor performances facing these scenarios. Note that, when the input image just contains homogeneous background plus detector noise (\autoref{fig:original6}),  since there is no structural texture, the performance of the AAGD algorithm is close to the proposed algorithm.

Background suppression factor (BSF) and signal to clutter ratio (SCR) which are defined as follows are used as quantitative measures:
\begin{equation}
BSF=\frac{\sigma_{in}}{\sigma_{out}}, \quad \quad \quad SCR=\frac{f_T-f_b}{\sigma_b} 
\label{eq:refname12}
\end{equation}
where $f_T$,  $f_b$, $\sigma_b$,  $\sigma_{in}$, and  $\sigma_{out}$ denote average value of target pixels, mean of background region, standard deviation of local background, standard deviation of non-target area in original image, and standard deviation of non-target area in filtered image, respectively. A higher BSF value indicates better background clutter rejection and the more SCR the more possibility to hit the true target. These values are reported in \autoref{tab:scrgmu} and \autoref{tab:bsf} for the test images. As reported in the tables, it is clear that the ADMD algorithm strongly outperforms other methods from both SCR and BSF points of view.

While BSF and SCR are pre-thresholding metrics, false-alarm rate ($P_{fa}$) is used to evaluate post-thresholding clutter rejection ability
\begin{equation}
 P_{fa}=\frac{N_f}{N_w} 
 \label{eq:refname17}
\end{equation}
where $N_f$ and $N_w$ are the number of false alarms and number of all pixels in the whole image, respectively \cite{moradi2016scale}. False-alarm rate versus global threshold level curves for the test images are shown in \autoref{fig:roc_test}. As depicted in the figure, the proposed algorithm has the lowest false-alarm rate compared to the other algorithms. This means that the proposed algorithm effectively suppresses background clutter.
\subsection{Execution time analysis}
In this section, the run-time analysis of the different algorithms is provided to demonstrate the computational complexity of each algorithm. All the algorithms are implemented using OpenCV C++ libraries. The full specifications of the implementation environment are reported in \autoref{tab:spec_runtime}. Each detection algorithm is executed 100 times and the average execution time for the multi-scale (except MS-LoG algorithm which utilizes 12 scales \cite{kim2012scale}, the other multi-scale algorithms consist of four identical scales)  as well as the single-scale versions of the algorithms are reported in \autoref{tab:ms_exe_time} and \autoref{tab:ss_exe_time}, respectively. As reported in the \autoref{tab:ms_exe_time}, the efficient implementation of the ADMD algorithm (ADMD$_{\text{eff}}$) is almost 27 times faster than the original ADMD. Since the whole $288\times 110000$ panorama image is captured every 1.5 seconds, none of these multi-scale algorithms can  operate in real-time using the reported hardware (\autoref{tab:spec_runtime}). GPU implementation will be a good choice for the real-time implementation of these algorithms. \autoref{tab:ss_exe_time} shows the average execution time for single-scale algorithms. As reported in the table, TopHat, AAGD and ADMD$_{\text{eff}}$ algorithms can be implemented in real-time using the reported hardware. Also, it can be seen that the efficient ADMD$_{\text{eff}}$ algorithm is significantly faster than the original ADMD algorithm.

\begin{table}[!b]
	\centering
	\caption{ The full specifications of the implementation environment}
	\begin{tabular}{|c|c|}
		\hline
	Operating System	 & Linux (Ubuntu 18.04, X64) \\
		\hline
		Linux kernel version &  4.15.0-42-generic  \\
		\hline
		OpenCV version  &  3.4.3 \\
		\hline
		Compiler &  GCC 7.3.0 \\
				\hline
		Size of the test image &   288$\times$5600 \\
		\hline
				data depth and type & single channel 32 bit floating point \\
				\hline
		CPU &  Intel CORE i7-3520M  @ 2.90GHz \\
		\hline
		Memory &   8GB DDR3 @ 1600MHz \\
		\hline
		\end{tabular}
	\label{tab:spec_runtime}
\end{table}
\begin{table}[!t]
	\centering
	\caption{ The average execution time for \textbf{\emph{multi-scale}} algorithms. ADMD$_{\text{eff}}$ denotes efficient implementation of the ADMD algorithm.}
	\begin{tabular}{|c|c|}
		\hline
	\textbf{Detection method}	 & \textbf{ execution time (mS)} \\
		\hline
		AAGD &  145.323  \\
		\hline
		LCM  &  4555.747 \\
		\hline
		LoG &  213.667 \\
				\hline
		PCM &   3463.643 \\
		\hline
		ADMD&4612.893 \\
				\hline
		ADMD$_{\text{eff}}$&  171.893 \\
		\hline
		\end{tabular}
	\label{tab:ms_exe_time}
\end{table}
\begin{table}[!t]
	\centering
	\caption{ The average execution time for \textbf{\emph{single-scale}} algorithms. ADMD$_{\text{eff}}$ denotes efficient implementation of the ADMD algorithm.}
	\begin{tabular}{|c|c|}
		\hline
	\textbf{Detection method}	 & \textbf{execution time (mS)} \\
		\hline
		AAGD, $7\times 7$ target window &  40.622  \\
		\hline
		LCM, $7\times 7$ cell size &  1111.990 \\
		\hline
		TopHat, $7\times 7$ flat SE &  4.536 \\
				\hline
		PCM, $7\times 7$ cell size &  1040.885 \\
		\hline
		ADMD, $9\times 9$ cell size&1195.462 \\
				\hline
						ADMD, $7\times 7$ cell size&1125.827 \\
				\hline
						ADMD, $5\times 5$ cell size&1109.403 \\
				\hline
						ADMD$_{\text{eff}}$, $9\times 9$ cell size&63.002 \\
				\hline
						ADMD$_{\text{eff}}$, $7\times 7$ cell size&46.405 \\
				\hline
		ADMD$_{\text{eff}}$, $5\times 5$ cell size&  32.049 \\
		\hline
		\end{tabular}
	\label{tab:ss_exe_time}
\end{table}
\section{Conclusion}
Infrared small target detection is of great importance in a wide range of applications such as surveillance, remote sensing, security, and astronomy. Despite the high attention paid to this field, the lack of robust and effective target detection method is tangible. To develop a robust small target detection algorithm, in this paper, a directional approach is constructed for better background suppression. Since the small targets have positive local contrast in all directions, in the proposed method, after calculating the directional cell-based contrast for all eight directions, the minimum contrast is used as the output value. 
For a small target (which is an isotropic object in the image), all the directional contrasts have almost equal values. Thus, the minimum selection strategy does not affect the target enhancement ability. However, structural background shows different responses along the different direction. For instance, a sharp edge has zero contrast along the edge direction. Therefore, the minimum selection strategy will eliminate structural backgrounds. It may seem that the noise suppression ability decreases using the proposed approach because of that the local background cell in the proposed method is smaller than the local background window in AAGD algorithm. However, using the minimum selection strategy in ADMD  suppresses background noise power much more compared to AAGD. Using SCR and BSF  as pre-thresholding metrics shows that the ADMD algorithm outperforms the other baseline algorithms. Also, by constructing the false-alarm rate versus  threshold values curve for the different algorithms, the advantage of the ADMD  over AAGD algorithm is demonstrated in post-thresholding situation. Compared to the AAGD algorithm, the ADMD utilizes eight different directional contrasts which in turn increases the execution time of the ADMD algorithm. To address this problem, an efficient implementation of the ADMD algorithm is proposed through the reformulation of the algorithm. The improved implementation just utilizes an extra morphological dilation with a specific structural element. The improved implementation is  almost 27 times faster than the original ADMD. All the proposed as well as the baseline algorithms are implemented using OpenCV C++ libraries. While the multi-scale version of the both AAGD and ADMD may not be implemented in real-time on old machines, the single-scale implementation of both algorithms is fast enough for practical applications.

\bibliographystyle{elsarticle-num}
\bibliography{master}
\end{document}